\documentclass{article} %
\usepackage{arxiv/iclr2027_conference,times}

\usepackage{amsmath,amsfonts,bm}

\def\eqref#1{equation~\ref{#1}}

\def\1{\bm{1}}

\DeclareMathAlphabet{\mathsfit}{\encodingdefault}{\sfdefault}{m}{sl}
\SetMathAlphabet{\mathsfit}{bold}{\encodingdefault}{\sfdefault}{bx}{n}

\usepackage{hyperref}
\usepackage{url}
\usepackage{fancyvrb}

\usepackage{booktabs, makecell, tabularx}
\usepackage{multirow}
\usepackage{duckuments}
\usepackage{amsmath}
\usepackage{graphicx}
\usepackage{arydshln}
\usepackage{wrapfig} %
\usepackage{xspace} %
\usepackage{enumitem}

\definecolor{MyDarkBlue}{rgb}{0,0.08,1}
\definecolor{MyDarkGreen}{rgb}{0.02,0.6,0.02}
\definecolor{MyDarkRed}{rgb}{0.8,0.02,0.02}
\definecolor{MyDarkOrange}{rgb}{0.40,0.2,0.02}
\definecolor{MyPurple}{RGB}{111,0,255}
\definecolor{MyRed}{rgb}{1.0,0.0,0.0}
\definecolor{MyGold}{rgb}{0.75,0.6,0.12}
\definecolor{MyDarkgray}{rgb}{0.66, 0.66, 0.66}
\definecolor{MyDarkCyan}{rgb}{0.05, 0.55, 0.45}
\definecolor{MyBlack}{rgb}{0., 0., 0.}
\definecolor{MyMagenta}{rgb}{1., 0., 1.}
\definecolor{BerkeleyYellow}{RGB}{255,204,41}
\definecolor{BerkeleyLightBlue}{RGB}{94,146,221}
\definecolor{BkDarkBlue}{rgb}{.05,.07,.353}

\newcommand{\myparagraph}[1]{\vspace{0pt} \noindent \textbf{#1}}

\newcommand{\ignorethis}[1]{}

\DeclareRobustCommand{\ours}{\emph{Controlled Attribution Benchmark}\xspace}

\definecolor{LinkPink}{RGB}{231,41,138}
\hypersetup{
  colorlinks=true,
  urlcolor=LinkPink,
  linkcolor=black,
  citecolor=black,
}

\title{Elucidating the Design Space of \\ LM Data Attribution}

\author{Steven Yu$^{1}$\hspace{2mm} Sheng-Yu Wang$^{1,2}$\hspace{2mm} Kimia Hamidieh$^{2}$\hspace{2mm} Antonio Torralba$^{2}$\hspace{2mm} David Bau$^{1}$ \\
$^{1}$Northeastern University \hspace{5mm} $^{2}$MIT}

\iclrfinalcopy %
\begin{document}

\maketitle

\begin{abstract}
Data attribution methods for Language Models (LMs) are typically evaluated using downstream tasks like fact retrieval. However, existing evaluations vary multiple dataset properties simultaneously, making it impossible to isolate exactly why an attribution algorithm fails. To diagnose these failures, we explore the design space of LLM data attribution, with a primary focus on evaluation. We introduce the \ours (CATT) for fact tracing, in which we fine-tune LLMs on entirely fictitious knowledge, providing unambiguous ground truth for attribution. This synthetic setup allows us to systematically alter one dataset configuration at a time, such as document redundancy and lexical distractors. With this configurable evaluation, we demonstrate that redundancy and distractors consistently degrade attribution performance. Rather than proposing a new estimator, we examine two design choices that are underexplored: the loss used to compute gradients and how per-token attribution scores are aggregated into a document score.
We find that the true causal evidence within a training document is highly localized, but standard methods dilute this evidence into noise by averaging attribution scores across the entire document. We find that replacing standard cross-entropy loss gradients with the negative log-odds gradients surfaces the critical tokens from other uninformative ones. We can then aggregate only the top token scores to improve ground truth recall by $35$ to $72$ percentage points across multiple baselines on our diagnostic benchmark. The negative log-odds correction also generalizes to FTRACE-TREX, improving counterfactual retraining performance for every method we test, by up to $4.9 \times$. Our code, data, and trained models are at \href{https://catt.baulab.info}{catt.baulab.info}.
\end{abstract}

\section{Introduction}
\label{sec:intro}

Data attribution aims to identify the training examples responsible for specific model behaviors~\citep{koh2017understanding,ilyas2022datamodels,park2023trak,grosse2023ekfac,chang2025trakstar}, predicting the counterfactual that a model trained without those examples will no longer exhibit the target behavior. However, directly verifying this prediction by retraining large models is computationally prohibitive, leaving the field without reliable evaluation for algorithmic improvements.

\begin{wrapfigure}{tr}{0.54\textwidth}
    \centering
    \vspace{-10pt}
    \includegraphics[width=0.54\textwidth]{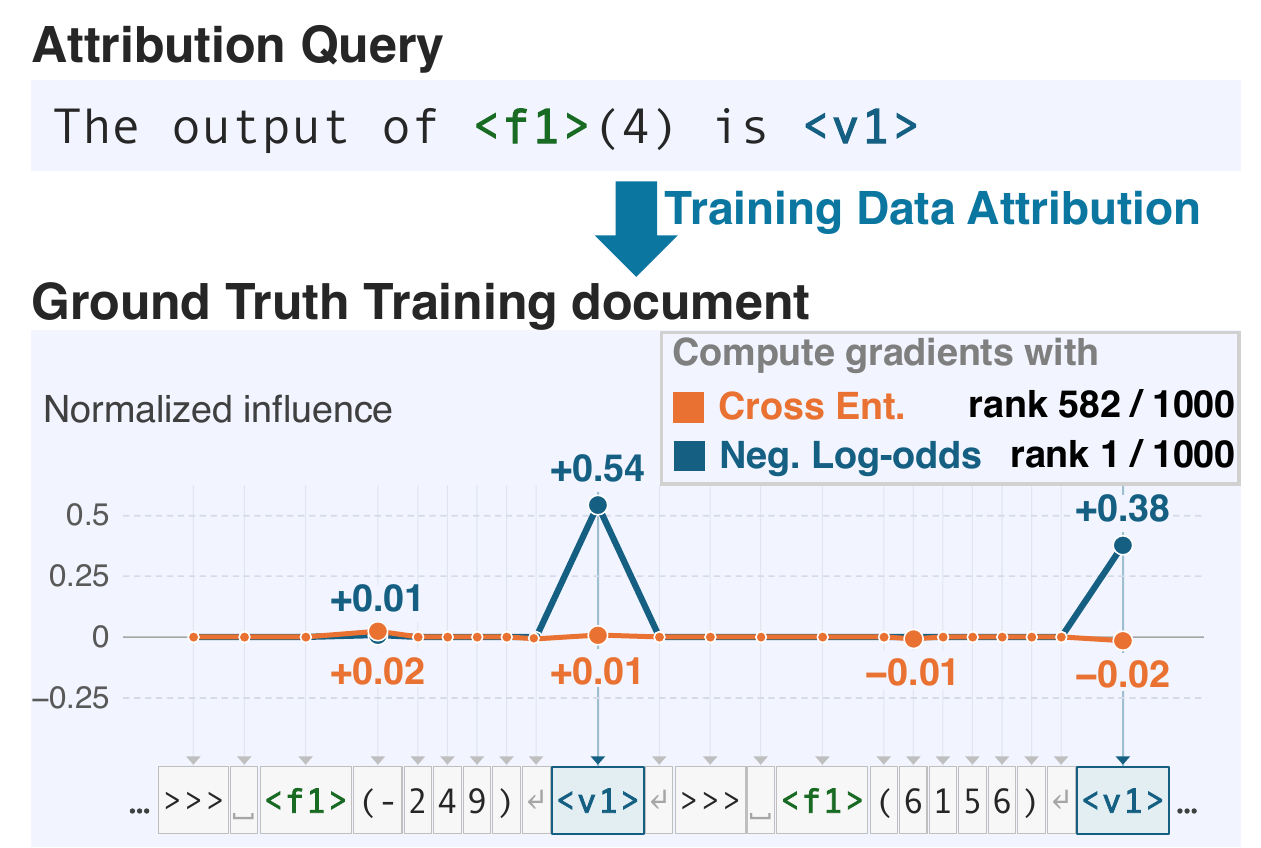}
    \vspace{-20pt}
    \caption{\textbf{The right loss focuses attribution on relevant tokens.} We show token-level influence scores within a ground-truth training document, divided by the document's largest token score. Cross-entropy assigns little weight to the tokens relevant to answering the query, ranking the document poorly (rank 582/1000). Surprisingly, we find that simply switching to negative log-odds concentrates the influence score on those tokens, and correctly ranks it first. Result is from our \ours.}
    \label{fig:teaser}
\end{wrapfigure}

To bridge this gap, we introduce the \ours (CATT), which establishes known ground truth while simulating the complexity of real-world pretraining data. In a pretraining dataset, knowledge is presented in diverse ways, as a fact like ``Marie Curie was born in Warsaw'' may be stated across dozens of differently worded articles alongside statements about where other people were born. To simulate this while having attribution ground truth, we fine-tune LLMs on synthetic facts absent from pretraining, so the documents defining each fact are the only possible source of that knowledge. Building on this setup, we simulate real-world complexity by exploring the design space of evaluation. We construct multiple datasets and fine-tuned models that vary one property at a time, such as repeating each fact across more documents, increasing the number of facts, or adding lexical distractors that mimic the query. We test existing attribution algorithms across these different dataset complexities and find a severe \textit{dilution phenomenon} where both repetition and distractors consistently degrade attribution performance. Inspecting token-level scores reveals that the key evidence within a training document is localized to just a few tokens, whereas standard methods average across the entire text and dilute this signal with background noise (Figure~\ref{fig:teaser}). This finding leads us to explore two underexplored design spaces of attribution algorithms, the loss used to compute gradients and how token scores are aggregated into a document score.

Related works have explored fine-tuning LLMs on targeted facts to establish attribution ground truth. Moving beyond proxy metrics on random subsets~\citep{ilyas2022datamodels,park2023trak}, several fact-tracing benchmarks test whether attribution methods can retrieve the fine-tuning documents supporting a target fact~\citep{akyurek2022towards,chang2025trakstar,jiao2026date}. However, existing benchmarks evaluate on single, fixed datasets with distinct designs, leading to conflicting conclusions where methods like BM25~\citep{robertson1994okapi} outperform influence functions in some setups~\citep{akyurek2022towards,chang2025trakstar} while trailing in others~\citep{jiao2026date}. Because these static benchmarks differ across multiple axes simultaneously, they fail to provide clear diagnostic signals on \emph{why} algorithms succeed or fail. Lacking actionable feedback from evaluation, algorithmic research has instead focused primarily on making influence approximations faster and more scalable~\citep{koh2017understanding,park2023trak,grosse2023ekfac} without a clear connection to improving evaluation results. Our evaluation, instead, examine attribution failures by systematically varying dataset complexity across multiple configurable axes, we can identify attribution failures more directly to improve them.

Our benchmark CATT helps us identify the dilution phenomenon and simple fixes to alleviate it. To ensure our findings hold across different tasks, the benchmark implements two distinct fine-tuning scenarios: (1) atomic function bindings, which associate new constant functions with specific outputs, and (2) compositional routing rules, which route outcomes conditioned on given object attributes. Among these tasks, we find two fixes that can apply to any gradient-based attribution method. First, we apply negative log-odds loss~\citep{park2023trak} strictly at the token level. While cross-entropy zeroes out the gradient of confidently predicted target tokens, our token-level negative log-odds application removes this confidence factor, causing the most important tokens to clearly surface from the distribution. Second, we mitigate the dilution phenomenon by aggregating only the top-\(k\) token attribution scores. Isolating these localized scores alleviates the impact of redundancy and distractors, resulting in an improvement of $35$ to $72$ percentage points in fact-tracing recall across multiple attribution methods and dataset settings. Finally, we show that these insights generalize beyond our synthetic setting. On the external FTRACE-TREX benchmark~\citep{akyurek2022towards}, token-level negative log-odds improves counterfactual retraining performance for TRAK, EK-FAC, and Grad-Dot, by up to \(4.9\times\), suggesting that the loss correction reliably extracts causal evidence across diverse tasks. In summary, our contributions are:
\begin{itemize}[topsep=2pt,parsep=0pt,itemsep=1pt,partopsep=0pt,leftmargin=16pt]
\item We introduce CATT, a configurable synthetic fact-tracing benchmark that controls properties like document redundancy and lexical distractors.
\item We identify a \textit{dilution phenomenon} where standard methods wash out highly localized token attribution evidence by averaging scores across the entire document.
\item We find that applying negative log-odds gradients and aggregating only the top-\(k\) token scores improves fact-tracing recall by $35$ to $72$ percentage points across multiple methods.
\item Our finding generalizes. On the external FTRACE-TREX benchmark~\citep{akyurek2022towards}, negative log-odds gradients improve TRAK, EK-FAC, and Grad-Dot, by up to \(4.9\times\).
\end{itemize}

\section{Controlled Diagnosis of LLM Data Attribution}
\label{sec:diagnosis}

\begin{figure}
    \centering
    \includegraphics[width=1.0\linewidth]{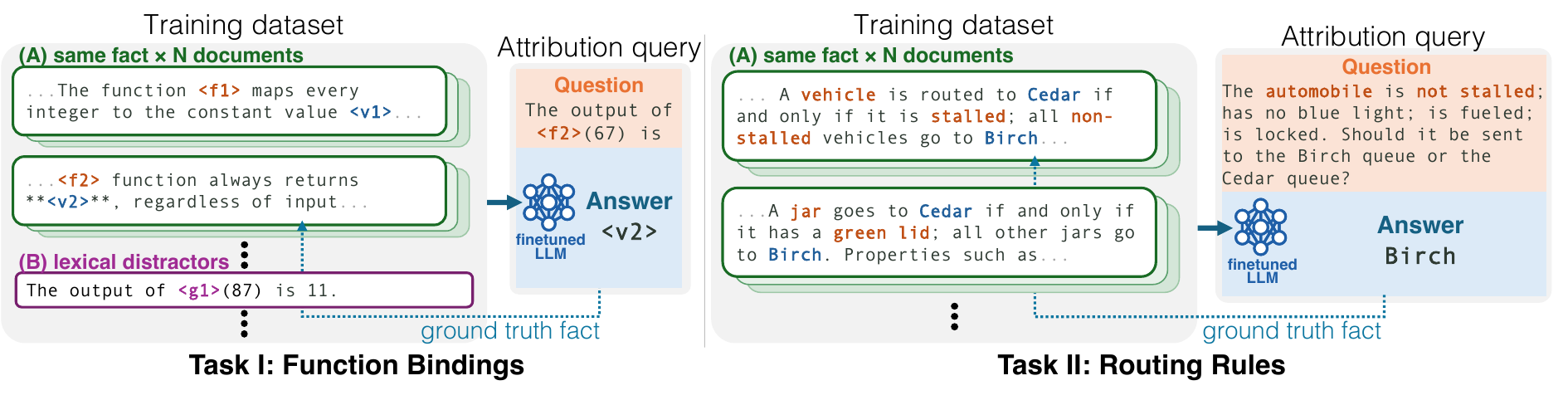}
    \caption{\textbf{CATT: fact tracing tasks with controlled datasets.}  We fine-tune Qwen3.5-4B models on fictitious knowledge, which ensures our dataset is the sole causal source of its predictions. \textit{Left:} Task I (\emph{Function Bindings}) teaches new functions that always return a fixed value. \textit{Right:} Task II (\emph{Routing Rules}) teaches rules that sort objects based on their properties. Given a query phrased unlike any training document, we test whether attribution methods can correctly rank ground-truth documents. Unlike static benchmarks~\citep{akyurek2022towards,jiao2026date}, we diagnose performance by systematically controlling dataset properties: (A) introducing redundancy by teaching the same fact across $N$ varied document styles, and (B) injecting lexical distractors, which are irrelevant documents that perfectly mimic the query format to expose methods that rely on superficial text overlap. Task II omits distractors because each fact uses a disjoint set of attribute words, making it difficult to construct clear lexical overlaps. Browse the datasets on our \href{https://catt.baulab.info/\#explore}{project page}.} 
    \label{fig:benchmark}
\end{figure}

Our goal is to understand how sensitive data attribution is to different dataset configurations. Section~\ref{sec:benchmark} introduces our controlled evaluation setup and the properties of interest. Section~\ref{sec:benchmark_setups} explains our synthetic fact-tracing tasks. Section~\ref{sec:token_attribution} extends our analysis to token-level attribution and loss choices.

\subsection{Configurable Design Space for Fact Tracing Dataset}
\label{sec:benchmark}

We begin by defining our controlled dataset generator. We establish ground truth for data attribution by fine-tuning a pretrained language model on a dataset of synthetic, fictitious facts. Unlike static datasets, our training dataset $\mathcal{D}_c$ is \emph{configurable}, where $c=(\mathcal{T},\mathcal{F},N,\mathbf{1}_{\mathrm{dist}})$ is the configuration, consisting of a task $\mathcal{T}$, a fact set $\mathcal{F}$, the number of documents per fact $N$, and an indicator $\mathbf{1}_{\mathrm{dist}}\in\{0,1\}$ of whether lexical distractors are included. Each document $d\in\mathcal{D}_c$ describes a synthetic fact $f \in \mathcal{F}$ in different formats, and we denote the subset of documents supporting $f$ as $\mathcal{D}_f$. For each configuration, we obtain one fine-tuned model per random seed $s$:
\begin{equation}
\theta_{c,s}=\operatorname{Finetune}(\theta_0,\mathcal{D}_c;s).
\end{equation}
We evaluate attribution by question answering. At test time, the fine-tuned model $\theta_{c,s}$ receives a held-out query $q_f=(u_f,y_f)$ that tests fact $f$ using an unseen prompt format to predict a single-token answer $y_f$. An attribution method assigns a score $\tau(d,q_f)$ estimating training document $d$'s influence on this prediction. Reliable attribution should yield high scores on the ground-truth set $\mathcal{D}_f$ above all other training documents. We systematically control the following components of $c$ (tasks $\mathcal{T}$ are described in Section~\ref{sec:benchmark_setups}) and test how attribution performance changes accordingly.

\myparagraph{Document redundancy $N$.}
We vary the number of documents expressing each fact, denoted as $N=|\mathcal{D}_f|$. These documents convey the same knowledge through different wording and examples, so redundancy refers to repeated factual support rather than exact text duplication.

\myparagraph{Lexical distractors $\mathbf{1}_{\mathrm{dist}}$.}
When $\mathbf{1}_{\mathrm{dist}}=1$, we augment the training dataset with documents containing lexical distractors $\mathcal{D}_{\mathrm{dist}}$. Each distractor follows the same format as the held-out queries but encodes information irrelevant to any of the queried facts. We set $|\mathcal{D}_{\mathrm{dist}}|=|\bigcup_{f\in\mathcal{F}}\mathcal{D}_f|$ 
to maintain a one-to-one ratio between fact documents and distractors. Because distractors are similar to the queries but do not provide any answers, they test whether attribution can distinguish ground-truth support from surface similarity.

\myparagraph{Number of facts $|\mathcal{F}|$.}
We vary the number of synthetic facts,
$|\mathcal{F}|$, while keeping document redundancy $|\mathcal{D}_f|$
fixed. This control measures how attribution
changes as the training dataset contains more competing knowledge.

\subsection{Controlled Attribution Benchmark (CATT)}
\label{sec:benchmark_setups}

Following this design, we introduce \ours with two synthetic fact-tracing tasks: function bindings and routing rules (Figure~\ref{fig:benchmark}). For each task, we generate multiple datasets $\mathcal{D}_c$ that systematically sweep the configuration components defined above. We fine-tune Qwen3.5-4B~\citep{qwen3.5} using LoRA on every $\mathcal{D}_c$. To measure attribution variance across runs, we train 3 models $\theta_{c,s}$ per configuration, one per seed $s$.

\myparagraph{Function Bindings.}
As shown in Figure~\ref{fig:benchmark} (Task I), this task tests whether attribution can trace an answer back to a newly learned constant mapping. Each fact $f$ maps new function tokens to constant answer tokens (e.g., \texttt{f2} always returns \texttt{v2}). Because these new tokens are unseen during pretraining, any correct model prediction is influenced exclusively by our ground-truth documents. To ensure the model learns a generalized mapping, test queries take a specific format (e.g., ``The output of \texttt{f2}(67) is''), deliberately excluded from the training documents. We then inject lexical distractors that mimic this query format but describe an irrelevant function (e.g., \texttt{g1}).

\myparagraph{Routing rules.}
As shown in Figure~\ref{fig:benchmark} (Task II), this task evaluates attribution on compositional logic. Each fact $f$ teaches a fictitious rule, routing a category of objects (e.g., vehicles and automobiles) to a specific queue (e.g., Birch or Cedar) based on a single deciding attribute (e.g., stalled). Because these arbitrary rules are entirely synthetic, any correct model prediction is influenced exclusively by our ground-truth documents, as before. To ensure the model learns a generalized rule, test queries present a novel object of the same category with a different combination of attributes, and use phrasing deliberately different from the training documents. Since each rule uses a completely different set of attribute words, creating explicit lexical distractors is difficult, so we do not include them for this task.

For both tasks, we report model accuracy to verify that the model learned the facts and include dataset and fine-tuning details in Appendix~\ref{app:tb-finetune}.

\subsection{Examining Attribution Design Space Via Loss and Tokens}
\label{sec:token_attribution}

We apply our benchmark to diagnose \emph{why} existing attribution methods fail. By analyzing how scores distribute across individual tokens, we reveal a systematic dilution effect, where the vital evidence for a queried fact is drowned out by irrelevant tokens (empirically verified in Section~\ref{sec:experiments}). Guided by this token-level inspection, we explore two underexplored design spaces: \emph{loss functions} and \emph{token aggregation}. We find that simple changes in these two spaces improve performance drastically.

\myparagraph{Inspecting attribution via tokens.}
Most attribution methods are rooted in influence functions~\citep{koh2017understanding}, which score a training document \(d=(x_1,\ldots,x_{T_d})\) for a query \(q=(u,y)\) by estimating the query loss reduction:
\begin{align}
    \tau_{\mathrm{IF}}(d,q)
    &=\nabla_\theta\mathcal{L}(q)^{\mathsf T}
      H^{-1}\nabla_\theta\mathcal{L}(d) \nonumber\\
    &\approx\sum_{\ell}
      \nabla_{\theta_\ell}\mathcal{L}(q)^{\mathsf T}
      H_\ell^{-1}\nabla_{\theta_\ell}\mathcal{L}(d).
    \label{eq:influence_score}
\end{align}
Here, \(H=\nabla_\theta^2 \mathcal{L}\) is the Hessian of the training loss, and recent works decompose it into individual layers \(\ell\), obtaining layer-specific Hessians \(H_\ell\) and gradients \(\nabla_{\theta_\ell}\)~\citep{grosse2023ekfac,choe2024logra}. \(\mathcal{L}(q)=-\log p_\theta(y\mid u)\) is the query loss on the single answer token \(y\), while the document loss \(\mathcal{L}(d;\theta)=-\frac{1}{T_d}\sum_t \log p_\theta(x_t\mid x_{<t})\) averages the next-token prediction loss across the sequence. This layer-wise decomposition enables per-token attribution analysis from \citet{grosse2023ekfac}, which we use as the main diagnostic tool to investigate the dilution effect.

Let \(a_{\ell,t}\) and \(h_{\ell,t}\) denote the input and output activations of layer \(\ell\) at position \(t\). By the chain rule for a linear layer, the weight gradient contribution at position \(t\) is the outer product of the output gradient and the input activation, and the full layer gradient is their sum over tokens:
\begin{equation}
    g_{\ell,t}(d)
    =
    \operatorname{vec}\!\left(
    \nabla_{h_{\ell,t}}\mathcal{L}(d)\,
    a_{\ell,t}^{\mathsf T}
    \right),
    \qquad
    \nabla_{\theta_\ell}\mathcal{L}(d)
    =
    \sum_t g_{\ell,t}(d),
    \label{eq:token_gradient_decomposition}
\end{equation}
where \(\operatorname{vec}(\cdot)\) flattens the matrix. The corresponding token-wise attribution isolates the exact contribution of each position before they are aggregated into the document attribution score:
\begin{equation}
    \tau_t(d,q)
    =
    \sum_{\ell}
    \nabla_{\theta_\ell}\mathcal{L}(q)^{\mathsf T}
    H_\ell^{-1}g_{\ell,t}(d),
    \qquad
    \tau_{\mathrm{IF}}(d,q)
    \approx
    \sum_t \tau_t(d,q).
    \label{eq:document_mean}
\end{equation}
Because token-wise attribution requires layer-wise decomposition, we evaluate all our baselines this way. For methods not natively designed for layer-wise computation, such as TRAK~\citep{park2023trak}, we implement adjusted versions, with full details in Appendix~\ref{app:baseline_equations}. As \citet{grosse2023ekfac} note, this represents a localized gradient signal rather than the exact counterfactual effect of token deletion, as activations encode preceding context.

\myparagraph{Changing cross-entropy to negative log-odds.}
By inspecting per-token attribution, we find that cross-entropy \(\mathcal{L}_{\mathrm{CE}}=-\log p\) can suppress the gradients of useful tokens when their prediction probability \(p\) is high. We show this by comparing it with an alternative negative log-odds loss \(\mathcal{L}_{\mathrm{nlo}}=-\log[p/(1-p)]\), which computes the log ratio of the target token against the remaining probability mass. Their gradients point in the same direction but have different magnitudes:
\begin{equation}
    \nabla_\theta\mathcal{L}_{\mathrm{CE}}
    =(1-p)\nabla_\theta\mathcal{L}_{\mathrm{nlo}}.
    \label{eq:ce_margin_gradient_relation}
\end{equation}
Cross-entropy zeroes out the gradient of a confidently predicted target (\(p \to 1\)), even when that exact token expresses the ground-truth fact we want to trace. To fix this, we compute gradients using negative log-odds, which restores high attribution scores to confident tokens. Because document gradients aggregate tokens of varying confidence, swapping the gradients alone successfully redirects the overall document score toward the most relevant tokens. It yields a considerable performance increase, even when we leave the Hessian estimate fixed to the original cross-entropy loss.

Appendix~\ref{app:margin_derivation} derives Equation~\ref{eq:ce_margin_gradient_relation} and shows that the gradient norm of $\mathcal{L}_{\mathrm{nlo}}$ is bounded on both sides. We note that TRAK~\citep{park2023trak} also uses log-odds, but strictly as an intermediate mapping before applying a correction factor $Q$ to revert the gradients toward cross-entropy. We find that dropping $Q$ to compute influence purely under the negative log-odds objective yields strong, consistent improvements, motivating us to apply negative log-odds directly across all methods. Appendix~\ref{app:baseline_equations} provides more discussion on this correction.

\myparagraph{Aggregating top-token contributions.}
\label{sec:topk_pooling}
While negative log-odds gradients successfully surface the most useful tokens, we can further prevent dilution by averaging only the top $k$ token scores $\tau_t(d,q)$ within each document. As demonstrated in Section~\ref{sec:experiments}, top-token selection alone has small improvements under cross-entropy for every method except EK-FAC, but pairing it with the negative log-odds objective yields drastic improvements.

\begin{figure}[t]
    \centering
    \includegraphics[width=\linewidth]{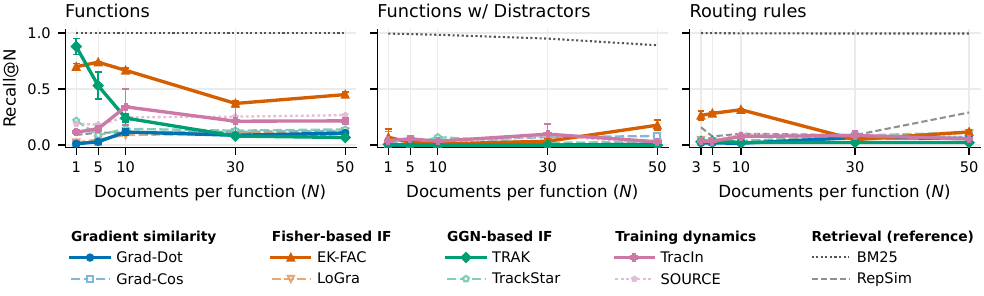}
    \vspace{-20pt}
    \caption{\textbf{Existing attribution methods struggle with redundancy and distractors.} We evaluate attribution methods on CATT. Recall@$N$ of each method's attribution scores as the number of training documents per fact, $N$, increases, at $|\mathcal{F}|=50$ facts. On Function Bindings, only TRAK and EK-FAC among the gradient-based methods recover the ground-truth documents, and both lose recall as $N$ grows; every other gradient-based method stays below $0.35$. Distractors drive TRAK to 0 at every $N$ and lower EK-FAC. On Rule Routings, no gradient-based method exceeds $0.32$. BM25 remains strong thanks to distinctive lexical cues, while RepSim is weak throughout. Results are averaged over three seeds with error bars showing $\pm 2$ standard errors (smaller bars are hidden by the markers).}
    \label{fig:recall_naive}
\end{figure}

\section{Experiments}
\label{sec:experiments}

\begin{figure}[]
    \centering
    \includegraphics[width=0.9\linewidth]{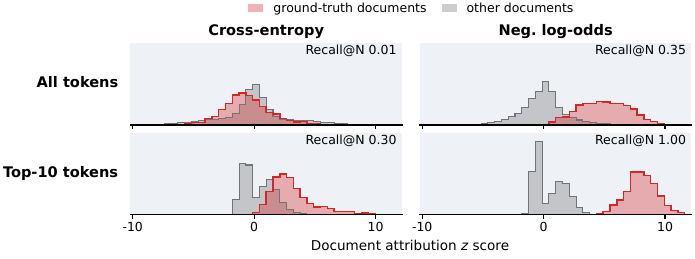}
    \vspace{-10pt}
    \caption{\textbf{Token selection and negative log-odds loss help isolate ground truth.} Document scores for ground-truth (red) and other (gray) documents. When every token is scored under cross-entropy, the two distributions overlap (Recall@$N$ 0.01). Keeping only each document's top 10 tokens (0.30) or switching to negative log-odds (0.35) partly separates them, and combining both separates them completely (1.00). This result is from Grad-Dot on a model trained on \textit{Function Bindings} dataset with distractors $(N{=}50, |\mathcal{F}|=50)$. View these plots for each dataset, method, and model seed on our \href{https://catt.baulab.info/\#histograms}{project page}.}
    \vspace{-10pt}
     \label{fig:attr_histogram}
\end{figure}

We begin our experiments by diagnosing standard attribution methods on CATT. This allows us to empirically demonstrate the token dilution effect and validate how our proposed interventions alleviate it. Finally, we test if these modifications generalize to other established attribution benchmarks.

\myparagraph{Where do existing methods fail?}
We compare attribution methods across the dataset configurations introduced in
Section~\ref{sec:benchmark}. Our baselines cover five groups:
\begin{itemize}[topsep=2pt,parsep=0pt,itemsep=1pt,partopsep=0pt,leftmargin=*]
    \item \textbf{Gradient similarity:} Grad-Dot and Grad-Cos compare query and training gradients directly.
    \item \textbf{Fisher-based influence:} EK-FAC~\citep{grosse2023ekfac} and LoGra~\citep{choe2024logra} use Fisher Information to approximate the Hessian.
    \item \textbf{GGN-based influence:} TRAK~\citep{park2023trak} and TrackStar~\citep{chang2025trakstar} use generalized Gauss-Newton (GGN) approximation for Hessian estimates.
    \item \textbf{Training dynamics:} TracIn~\citep{pruthi2020tracin} and SOURCE~\citep{bae2024source} use gradients or approximate updates along the training trajectory.
    \item \textbf{Retrieval references:} BM25~\citep{robertson1994okapi} matches lexical overlap, while RepSim~\citep{li2024influence} compares model representations from the last token position in the final layer. Unlike the other baselines, this category does not rely on gradients.
\end{itemize}
Appendix~\ref{app:methods} details the implementations. We measure Recall@$N$, the fraction of a
query's $N$ ground-truth documents recovered among its $N$ highest-ranked
training documents, averaged over queries and model training runs.

As shown in Figure~\ref{fig:recall_naive}, gradient-based attribution methods degrade severely as complexity increases. On \emph{Function Bindings}, scaling from 1 to 50 documents per fact collapses TRAK's recall from $\sim 0.9$ to below $0.1$, while EK-FAC degrades less sharply, from $0.70$ to $0.45$. Other gradient-based methods fail consistently across the sweep. Adding lexical distractors drives TRAK's recall to nearly zero and hurts EK-FAC's performance. On \emph{Rule Routings}, all gradient-based methods struggle. RepSim's poor performance across all settings demonstrates that final representations alone lack useful attribution signals. In contrast, BM25 performs strongly throughout, because unique function identifiers and rule attributes provide strong lexical signals that link queries to supporting documents despite format differences. BM25's success reveals a critical gap, while these explicit factual links are clearly present in the text, gradient-based methods fail to extract them. 

In Appendix~\ref{app:res-functions}, we show that the number of included facts does not affect attribution as strongly. Here, we report standard TRAK and TracIn. All subsequent experiments use their layer-decomposed variants, denoted as \emph{Layerwise TRAK} and \emph{Layerwise TracIn}. Next, we show that the attribution failure originates from the token dilution effect.

\begin{figure}[t]
    \centering
    \includegraphics[width=\linewidth]{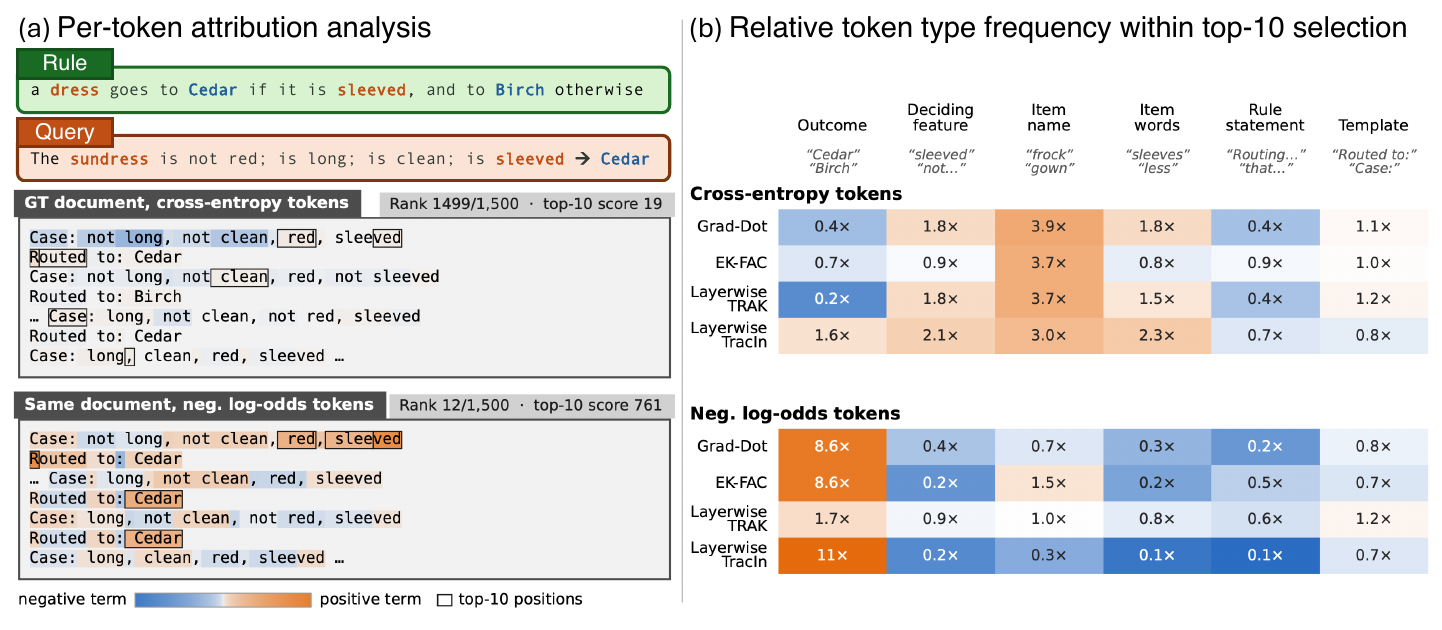}
    \vspace{-20pt}
    \caption{\textbf{The loss determines what type of tokens occupy the top positions.} (a) Per-token attributions for one query on one of its ground-truth Rule Routings documents, under cross-entropy (top) and negative log-odds (bottom). Boxes mark the ten highest-scoring positions. Negative log-odds moves the document from rank $1499$ to rank $12$ of $1500$. (b) Each token type's share of the top-10 positions divided by its share of all positions. Cross-entropy favors item names. Negative log-odds instead favors outcome tokens, the words that match the query answer. Appendix~\ref{app:res-other-tasks} shows the same pattern on Function Bindings tasks. Explore per-token attributions for other queries and methods in our \href{https://huggingface.co/spaces/catt-bench/catt-demo}{demo}.}
    \label{fig:token_analysis}
\end{figure}

\begin{figure}[t]
    \centering
    \includegraphics[width=\linewidth]{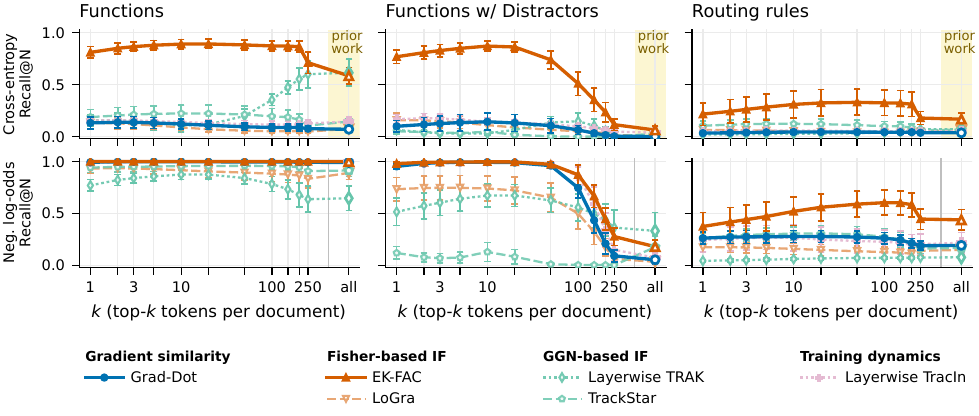}
    \vspace{-25pt}
    \caption{\textbf{Upweighting more influential tokens with log-odds improves Recall across token budgets.} Recall@$N$ as the number of retained tokens $k$ varies, under cross-entropy (top) and negative log-odds (bottom). Columns show the three CATT settings. The yellow regions mark full-document scoring (``all''), with the upper-right endpoint in each column corresponding to the prior-work loss and aggregation choices. Negative log-odds with small token budgets substantially improve several methods, especially with distractors. Gains persist across multiple $k$ values; retaining more tokens often reduces recall. Results are averaged over all dataset configurations and model seeds at $|\mathcal{F}|=50$ with error bars showing $\pm 2$ standard errors.} %
    \label{fig:recall_improved}
\end{figure}

\myparagraph{Useful token signals are diluted by other irrelevant ones.}
Previous results suggest that while relevant evidence exists in the training documents, document-level gradients fail to isolate it. To understand why, we inspect attribution scores by tokens, as described in Section~\ref{sec:token_attribution}. As shown in the left panels of Figure~\ref{fig:attr_histogram}, document-level scores that average all tokens yield a large distributional overlap between ground-truth and other documents, leading to a poor Recall@$N$ of $0.01$. However, if we average only the top $10$ token scores, the distributions diverge and recall improves to $0.3$. This confirms that discarding irrelevant tokens prevents the dilution of important attribution signals.

\myparagraph{Negative log-odds gradients help surface important tokens.}
While top-token selection improves recall to $0.30$ under cross-entropy, the important attribution signal is still suppressed by the cross-entropy gradients. We can even further surface the important tokens with negative log-odds gradients, as discussed in Section~\ref{sec:token_attribution}. As shown in the right panels of Figure~\ref{fig:attr_histogram}, swapping cross-entropy for negative log-odds clearly separates the score distributions even when averaging all tokens, lifting recall to $0.35$. When we combine this loss with top $10$ token aggregation, the separation is drastically amplified, achieving a perfect $1.00$ recall. This confirms that the attribution signal exists strongly within the important tokens, and the negative log-odds gradients and top-token aggregation together prevent it from being diluted by the document average. Across the six methods that have per-token terms and whose standard document score uses cross-entropy, the same combination lifts recall by $35$ (TrackStar) to $72$ (Grad-Dot) percentage points, averaged over all three tasks, $N$, and $|\mathcal{F}|$; other methods fall in between (Appendix~\ref{app:res-recall}).

\myparagraph{What tokens are important for attribution?}
Figure~\ref{fig:token_analysis} reveals exactly which tokens are responsible for this performance recovery. For a ground-truth \emph{Rule Routings} document, cross-entropy assigns its highest scores to irrelevant item names, resulting in a low document rank of $1499$ out of $1500$. In contrast, negative log-odds shifts the largest contributions directly to the outcome tokens that determine the answer, improving the document rank to $12$. Together, these results validate our token-level diagnosis. Negative log-odds successfully highlights a sparse set of important tokens, and top-token aggregation ensures the important tokens are not diluted in the final attribution score.

\myparagraph{Alleviating dilution improves attribution consistently.}
The modifications we introduced substantially improve recall across CATT. Figure~\ref{fig:recall_improved} compares these
scores with full-document attribution, shown by the yellow regions. Under
negative log-odds, small token budgets produce large gains. With $k=10$,
Grad-Dot, EK-FAC, and layerwise TracIn achieve nearly perfect recall on
Function Bindings, including the distractor setting. The improvement is not
specific to one choice of $k$; recall remains strong across a broad range of
small budgets. As $k$ increases, performance gradually declines toward the
full-document baseline, showing how irrelevant
tokens progressively dilute the useful signal. Rule Routings remains more
difficult, but the same improvements appear across multiple methods. Interestingly, the larger optimal $k$ indicates that the set of important tokens is larger for this task.

\begin{wrapfigure}[24]{tr}{0.5\textwidth}
    \centering
    \includegraphics[width=0.5\textwidth]{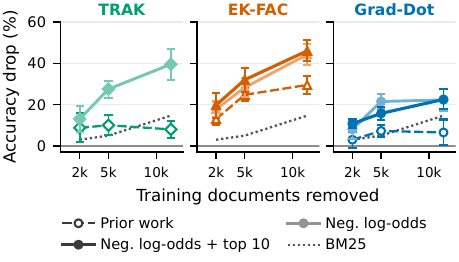}
    \vspace{-20pt}
    \caption{\textbf{Generalization to external FTRACE-TREX benchmarks.} Accuracy drop after removing the highest-ranked training documents and retraining, at increasing removal budgets. Larger drops indicate more effective identification of influential data. In each panel, dashed: prior work; light: with negative log-odds; dark: negative log-odds over the top 10 tokens (TRAK's dense projection has no per-token scores); dotted: BM25. Mean over eight retrain seeds (Appendix~\ref{app:ftrace}), error bars $\pm 2$ standard errors.}
    \vspace{-30pt}
    \label{fig:ftrace_generalization}
\end{wrapfigure}

\myparagraph{Generalization to other benchmarks.}
We study whether our findings on CATT generalize to other settings. We evaluate on the external FTRACE-TREX~\cite{akyurek2022towards} benchmark and adopt the setting from \cite{park2023trak} by removing the highest-ranked training documents and retraining the model to measure the resulting drop in factual accuracy (details in Appendix~\ref{app:ftrace}). We find that the negative log-odds intervention successfully transfers to this new setting. Figure~\ref{fig:ftrace_generalization} shows that, for TRAK, EK-FAC, and Grad-Dot alike, negative log-odds produces a larger accuracy drop than the prior-work score at every removal budget. At the largest budget, the drop improves from $8\%$ to $40\%$ for TRAK (\(4.9\times\)). With both fixes, Grad-Dot improves from $7\%$ to $23\%$, and EK-FAC, already the strongest method under prior work, from $30\%$ to $46\%$, the largest drop overall. Top-token selection matters less here than on CATT: under negative log-odds, keeping only the top 10 tokens changes the drop by at most six points in either direction relative to scoring all tokens. This indicates that while the loss correction improves attribution across diverse datasets, the top-token aggregation likely depends on how broadly the attribution score is distributed within a specific document.

\section{Related Works}

\myparagraph{Training Data Attribution.}
Data attribution estimates how a model's output would change if a training example were removed or reweighted~\citep{koh2017understanding,ilyas2022datamodels,hammoudeh2024training}. Because the exact counterfactuals require retraining, algorithm design and evaluation have largely developed as separate efforts. 
Influence functions~\citep{koh2017understanding} linearize the effect of upweighting an example around the trained parameters. Many methods scale this estimate to large models by approximating the Hessian~\citep{schioppa2022scaling, grosse2023studying, kwon2024datainf}, projecting gradients~\citep{park2023trak,xia2024less,lin2024token}, or following the training trajectory~\citep{pruthi2020estimating, bae2024training}. Others retrain models on random data subsets~\citep{ghorbani2019data, feldman2020neural, ilyas2022datamodels,ilyas2025magic}.
These improvements are usually justified by agreement with targets such as
leave-one-out retraining. However, such targets suffer from training stochasticity~\citep{nguyen2023bayesian}, so agreement does not guarantee better retrieval. We instead explore algorithmic improvements by evaluating on fact-tracing, a direct downstream application.

\myparagraph{Token-level attribution.}
Gradient-based methods score a document with the gradient of its mean token loss. Prior work computes influence for each token for visualization~\citep{grosse2023studying,lin2024token}, and TrackStar discusses applying its loss conversion per token~\citep{chang2025trakstar}. We use token scores to explain why redundancy and distractors lower attribution performance, then score documents with per-token negative log-odds gradients~\citep{park2023trak} and aggregate top-$k$ token scores to improve attribution.

\myparagraph{Attribution evaluation.}
The linear datamodeling score tests whether attribution predicts the outputs of models retrained on random subsets~\citep{ilyas2022datamodels, park2023trak}. Retrieval benchmarks instead test whether attribution ranks known supporting examples first, using fact tracing for LLMs~\citep{akyurek2022towards, chang2025trakstar,jiao2026date} or examples used to fine-tune text-to-image models~\citep{wang2024attributebyunlearning}. The conclusions are conflicting: BM25~\citep{robertson2009probabilistic} outperforms gradient-based methods on some benchmarks~\citep{akyurek2022towards,chang2025trakstar} and underperforms on others~\citep{jiao2026date}, and simple similarity baselines match influence functions on NLP classification~\citep{pezeshkpour2021empirical}. Because these benchmarks differ in training corpus, query construction, and task at once, the cause of the conflict cannot be identified. Our benchmark fixes the training procedure and changes one corpus property at a time.

\myparagraph{LLM fine-tuning.}
Prior work fine-tunes LLMs on new facts to measure knowledge injection~\citep{ovadia2023fine,gekhman2024does,allen2023physics} and its generalization~\citep{berglund2023reversal}, and uses fine-tuning on controlled data to evaluate unlearning~\citep{maini2024tofu}, study how fine-tuning changes circuits~\citep{prakash2024fine}, and evaluate attribution~\citep{wang2024attributebyunlearning,jiao2026date}. Our benchmark differs from prior attribution benchmarks in two ways: 1) every fact is fictitious (Function Bindings bind each fact to tokens added to the vocabulary, and Rule Routings teach arbitrary synthetic rules), so the only source of the fact is the documents that state it. 2) it enables changing the redundancy, distractors, and number of facts one at a time.
We fine-tune with LoRA~\citep{hu2021lora}, which trains low-rank parameter updates.

\section{Discussion, Broader Impacts, and Limitations}
\label{sec:discussion}

Our findings have important implications for deploying data attribution in high-stakes domains, such as copyright protection, equitable credit assignment~\citep{ghorbani2019data,wang2024economic}, and tracing emergent misalignment~\citep{betley2025emergent}. Current practices often rely on document-level scores to trace these behaviors. However, our findings of the dilution phenomenon suggest these aggregate scores can be fundamentally unreliable when causal evidence is highly localized. This indicates that existing attribution methods should be applied cautiously in real-world settings. Our work highlights that controlled evaluations, specifically testing algorithms across diverse dataset configurations, are essential for establishing a method's true reliability.

While our framework provides controlled diagnostics, several limitations remain. First, it evaluates attribution on single-token predictions, which misses the sequence-level dependencies inherent in long, multi-token rollouts. Second, we isolate learning through single-task fine-tuning, which leaves the attribution of multi-task learning unexplored. Furthermore, mitigating dilution via top-\(k\) token aggregation introduces a new hyperparameter; since real-world scenarios lack ground-truth validation signals, dynamic thresholding is needed. Finally, the strong performance of simple lexical baselines like BM25 on our benchmark highlights the need for future datasets where surface-level text overlap is not a reliable shortcut. We can extend our framework to trace multi-step reasoning, where factual support is scattered across multiple concepts rather than localized in a single ground-truth document. Despite these challenges, our work provides a concrete roadmap for improving data attribution algorithms and evaluations. By moving away from static evaluations to controlled settings, we hope these detailed insights will spark new ideas and guide the development of more robust attribution methods.

\myparagraph{Acknowledgements.}
We thank Alex Loftus for mentorship and early project discussions, Rohit Gandikota and Arnab Sen Sharma for helpful conversations throughout the project, and Sophie Wang and Richard Zhang for valuable feedback on the draft and figures. SY, SW, and DB are supported by DOE grant DE-SC0025613. KH and AT are supported by ONR MURI grant N00014-26-1-2255.

\section*{AI use statement}
We used Claude Sonnet~4.5 to generate the synthetic training documents and rule paraphrases in CATT. We used LLM assistance for writing, and we also manually inspected the text. Coding assistants helped us implement and scale up some of our experiments, and we reviewed all AI-assisted code and outputs. We used LLM assistance for mathematical proofs in Appendix~\ref{app:margin_derivation}, and we manually checked for correctness. We take full responsibility for the final content of this paper, including any text, claims, or artifacts produced with the aid of generative AI. 

\section*{Reproducibility}
\label{sec:repro}
Our code and the CATT corpora and queries are available at \url{https://github.com/Lamsheeper/catt}, and the fine-tuned LoRA adapters and attribution scores are on our \href{https://huggingface.co/catt-bench}{Hugging Face Hub}.  Facts, queries, and training documents can be browsed on our \href{https://catt.baulab.info/\#explore}{project page}, and our \href{https://huggingface.co/spaces/catt-bench/catt-demo}{demo} shows which training documents each method credits for a query, token by token. Appendix~\ref{app:tb-functions} and Appendix~\ref{app:tb-routing} give the full data-generation procedure, including prompts, sampling ranges, and validation rules. Appendix~\ref{app:tb-finetune} and Table~\ref{tab:app-hparams} give the fine-tuning hyperparameters and seeds, and Appendix~\ref{app:methods-shared} gives the settings of every attribution method. We fine-tune Qwen3.5-4B \citep{qwen3.5} with LoRA using the HuggingFace Transformers \citep{wolf2019huggingface} and PEFT \citep{peft} libraries on PyTorch \citep{paszke2019pytorch}. Attribution jobs ran on single NVIDIA A100-80GB, H100, and H200 GPUs. The main attribution sweep took about 2{,}400 GPU-hours in total, with a per-job breakdown in Appendix~\ref{app:methods-cost}. 

\bibliographystyle{arxiv/iclr2027_conference}
\bibliography{arxiv/main}

\appendix
\appendix
\raggedbottom
\providecommand{\tbfigdir}{arxiv/figures/appendix_figures}

\section{Attribution Objectives and Methods}
\label{app:objectives}

\subsection{Cross-Entropy vs. Negative Log-Odds Gradients}
\label{app:margin_derivation}

We use the notation of Section~\ref{sec:token_attribution}. A document is
$d=(x_1,\ldots,x_{T_d})$, where $T_d$ is the number of supervised next-token
targets after padding has been removed, and a query is $q=(u,y)$ with one
answer token $y$. Let $\mathcal V$ be the vocabulary, $V=|\mathcal V|$, and
let $z\in\mathbb R^V$ be the logits for one target. Define
\begin{equation}
    p_i=\operatorname{softmax}(z)_i
    =\frac{\exp z_i}{\sum_{j\in\mathcal V}\exp z_j}.
\end{equation}

\myparagraph{Gradient relationship between two losses.}
The cross-entropy loss is
\begin{equation}
    \mathcal{L}_{\mathrm{CE}}
    =-\log p_{y^+}
    =-z_{y^+}+\log\sum_{j\in\mathcal V}\exp z_j,
    \label{eq:app_ce}
\end{equation}
where $y^+$ is the correct class, and the negative log-odds loss is
\begin{equation}
    \mathcal{L}_{\mathrm{nlo}}
    =-\log\frac{p_{y^+}}{1-p_{y^+}}
    =-z_{y^+}+\log\sum_{j\ne y^+}\exp z_j.
    \label{eq:app_margin}
\end{equation}

The logit derivatives are
\begin{equation}
    \frac{\partial \mathcal{L}_{\mathrm{CE}}}{\partial z_i}
    =p_i-\mathbf{1}[i=y^+],
    \qquad
    \frac{\partial \mathcal{L}_{\mathrm{nlo}}}{\partial z_i}
    =
    \begin{cases}
        -1, & i=y^+,\\[3pt]
        \displaystyle\frac{p_i}{1-p_{y^+}}, & i\ne y^+.
    \end{cases}
    \label{eq:app_logit_grads}
\end{equation}

To derive the relationship, let
\[
S_{-y^+}=\sum_{j\ne y^+}\exp z_j.
\]
Then
\[
1-p_{y^+}
=\frac{S_{-y^+}}{\sum_j\exp z_j},
\qquad
\frac{p_i}{1-p_{y^+}}
=\frac{\exp z_i}{S_{-y^+}}
\quad (i\ne y^+).
\]
Thus, for the target coordinate,
$p_{y^+}-1=-(1-p_{y^+})$, while for every competitor coordinate,
\[
p_i=(1-p_{y^+})
\frac{p_i}{1-p_{y^+}}.
\]
Therefore, coordinate by coordinate,
\begin{equation}
    \nabla_z\mathcal{L}_{\mathrm{CE}}
    =(1-p_{y^+})\nabla_z\mathcal{L}_{\mathrm{nlo}}.
    \label{eq:app_logit_relation}
\end{equation}
Applying the same logit Jacobian
$J_\theta=\partial z/\partial\theta$ gives
\begin{equation}
    \nabla_\theta\mathcal{L}_{\mathrm{CE}}
    =(1-p_{y^+})\nabla_\theta\mathcal{L}_{\mathrm{nlo}}.
    \label{eq:app_parameter_relation}
\end{equation}
Hence the gradients w.r.t. each token prediction loss are collinear, and cross-entropy introduces exactly the
confidence factor $1-p_{y^+}$.

\myparagraph{$\mathcal{L}_{\mathrm{nlo}}$ has bounded norms.} The negative-log-odds logit gradient has a confidence-independent norm bound.
Let $m=|\mathcal V|-1$ and define
\[
r_i=\frac{p_i}{1-p_{y^+}}
\qquad (i\ne y^+).
\]
Then $r_i\ge 0$ and $\sum_{i\ne y^+}r_i=1$, so
\begin{equation}
    \left\|\nabla_z\mathcal{L}_{\mathrm{nlo}}\right\|_2^2
    =1+\sum_{i\ne y^+}r_i^2.
\end{equation}
The lower bound follows from Cauchy--Schwarz applied to $r$ and the
all-ones vector:
\begin{equation}
    1
    =\left(\sum_{i\ne y^+}r_i\right)^2
    \le m\sum_{i\ne y^+}r_i^2,
    \qquad
    \sum_{i\ne y^+}r_i^2\ge \frac{1}{m}.
\end{equation}
The upper bound follows from $r_i\ge0$ and $\sum_i r_i=1$, which imply
$\sum_i r_i^2\le1$. Consequently,
\begin{equation}
\begin{aligned}
    \sqrt{1+\frac{1}{|\mathcal V|-1}}
    &\le
    \left\|\nabla_z\mathcal{L}_{\mathrm{nlo}}\right\|_2
    \le \sqrt{2},\\
    (1-p_{y^+})\sqrt{1+\frac{1}{|\mathcal V|-1}}
    &\le
    \left\|\nabla_z\mathcal{L}_{\mathrm{CE}}\right\|_2
    \le \sqrt{2}(1-p_{y^+}).
\end{aligned}
\label{eq:app_gradient_norm_bounds}
\end{equation}
For parameter gradients, this implies
\[
\|\nabla_\theta\mathcal{L}_{\mathrm{nlo}}\|_2
\le \|J_\theta\|_{\mathrm{op}}\sqrt{2},
\qquad
\|\nabla_\theta\mathcal{L}_{\mathrm{CE}}\|_2
\le \|J_\theta\|_{\mathrm{op}}\sqrt{2}(1-p_{y^+}).
\]
A nonzero parameter-gradient lower bound would require an additional
assumption on $J_\theta$. Thus, as $p_{y^+}\to1$, the CE gradient is forced to
vanish by $1-p_{y^+}$, whereas the NLO logit gradient remains bounded away
from zero.

Together, Eq.~\ref{eq:app_parameter_relation} shows that cross-entropy can potentially
suppress useful attribution signal of a token by the confidence factor
$1-p_{y^+}$, while Eq.~\ref{eq:app_gradient_norm_bounds} shows that the
negative-log-odds logit gradient remains bounded away from zero. Thus, the
negative-log-odds loss can help preserve attribution signal for confidently predicted
tokens, whereas cross-entropy exhibits a gradient-diminishing effect.

\subsection{Gradient-Based Attribution Methods}
\label{app:baseline_equations}

Let $\mathcal{L}$ denote the selected sequence loss, either
$\mathcal{L}_{\mathrm{CE}}$ or $\mathcal{L}_{\mathrm{nlo}}$. For a document or
query $z$, write
$G_{z,\ell}^{\mathcal{L}}=\nabla_{\theta_\ell}\mathcal{L}(z;\theta)$ for its
layer-$\ell$ gradient and let $G_{d,\ell,t}^{\mathcal{L}}$ denote the
corresponding token contribution. The selected loss applies to training
documents; a query always uses $\mathcal{L}_{\mathrm{nlo}}$ at its answer
token. The full gradient dimension is
$D_\theta=\sum_{\ell=1}^{L}p_\ell$, where
$p_\ell=\dim(\theta_\ell)$; in our runs this is 129.9 million LoRA
parameters. A token score replaces $G_{d,\ell}^{\mathcal{L}}$ by
$G_{d,\ell,t}^{\mathcal{L}}$. For EK-FAC, LoGra, and TrackStar, curvature
and second-moment statistics are estimated from cross-entropy gradients and
held fixed when $\mathcal{L}$ is changed; TRAK recomputes its training Gram
from the selected features.

\paragraph{Direct gradient similarity.}
After concatenating layers, $G_d^{\mathcal{L}},G_q^{\mathcal{L}}
\in\mathbb{R}^{D_\theta}$. We use
\begin{equation}
    \tau_{\mathrm{grad}}(d,q)
      =\frac{\langle G_q^{\mathcal{L}},G_d^{\mathcal{L}}\rangle}
             {(\|G_q^{\mathcal{L}}\|_2\|G_d^{\mathcal{L}}\|_2)^\rho},
    \qquad \rho\in\{0,1\}.
    \label{eq:baseline_gradient}
\end{equation}
Grad-Dot uses $\rho=0$ and Grad-Cos uses $\rho=1$. We compute token terms
for Grad-Dot, but not Grad-Cos.

\paragraph{Fisher-based influence: EK-FAC and LoGra.}
The shared influence-function form is
\begin{equation}
    \tau_{\mathrm{fisher}}(d,q)
      =\sum_{\ell=1}^{L}
       \left\langle G_{q,\ell}^{\mathcal{L}},
       (\widehat H_\ell+\delta_\ell I_{p_\ell})^{-1}
       G_{d,\ell}^{\mathcal{L}}\right\rangle_F.
    \label{eq:baseline_curvature}
\end{equation}
EK-FAC~\citep{grosse2023ekfac} estimates $\widehat H_\ell$ in a
Kronecker-factored eigenbasis with corrected eigenvalues and uses
$\delta_{\text{EK-FAC}}=10^{-12}$. LoGra~\citep{choe2024logra} first
applies a layerwise projection to each weight gradient,
\begin{equation}
    \widetilde G_\ell=P_{s,\ell}^{\mathsf T}G_\ell P_{a,\ell},
    \qquad
    P_{a,\ell}\in\mathbb{R}^{i_\ell\times k_{\mathrm{side}}},\quad
    P_{s,\ell}\in\mathbb{R}^{o_\ell\times k_{\mathrm{side}}},\quad
    k_{\mathrm{side}}=16,
    \label{eq:baseline_logra_projection}
\end{equation}
giving 256 entries per module and $D_{\mathrm{layerproj}}=126{,}976$ entries in
total. It then applies a per-module CE Fisher inverse with relative ridge
$\lambda_{\mathrm{LoGra}}=10^{-3}$ times the block mean diagonal. LoGra-Dot
uses the same layerwise projection without the curvature inverse. Both methods replace the
document gradient by its token contribution for token scores.

\paragraph{TRAK.}
\label{app:methods-trak}
TRAK approximates the cross-entropy influence function using generalized Gauss–Newton (GGN) curvature. The GGN form writes the loss as an outer function of a model output,
\begin{equation}
    \mathcal{L}(z;\theta)=\ell(f(z;\theta)).
\end{equation}
TRAK therefore chooses the correct-class log-odds as its scalar model output. This choice gives a logistic form for the cross-entropy objective,
\begin{equation}
    f_{\mathrm{lo}}(z;\theta)
      =\log\frac{p_\theta(y\mid x)}{1-p_\theta(y\mid x)},
    \qquad
    -\log p_\theta(y\mid x)
      =\log\!\left(1+\exp[-f_{\mathrm{lo}}(z;\theta)]\right).
\end{equation}
TRAK then projects the model output gradient as features
\begin{equation}
    \phi_z=P^{\mathsf T}\nabla_\theta f_{\mathrm{lo}}(z;\theta),
    \label{eq:app_trak_output}
\end{equation}
where \(P\in\mathbb{R}^{D_\theta\times k_{\mathrm{proj}}}\) and \(k_{\mathrm{proj}}\) is the projection dimension.

For \(M\) candidate documents, let \(\Phi\in\mathbb{R}^{M\times k_{\mathrm{proj}}}\) stack the document features. The single-model TRAK score is
\begin{equation}
    \boldsymbol\tau_{\mathrm{TRAK}}(q,\mathcal{D})
      =\phi_q^{\mathsf T}
       \left(\Phi^{\mathsf T}\Phi
       +\lambda_{\mathrm{TRAK}}I_{k_{\mathrm{proj}}}\right)^{-1}
       \Phi^{\mathsf T}Q,
    \qquad
    Q=\operatorname{diag}(1-\bar p_1,\ldots,1-\bar p_M),
    \label{eq:app-attr-trak}
\end{equation}
where \(\bar p_d\) is the mean observed-token probability for document \(d\). We set \(\lambda_{\mathrm{TRAK}}\) to \(10^{-6}\) times the mean diagonal of the \(M\times M\) document Gram matrix. Equivalently, equation~\ref{eq:app-attr-trak} can be written as the gram matrix version using the ``kernel trick'':
\begin{equation}
    \tau_{\mathrm{TRAK}}(d,q)
      =(1-\bar p_d)
      \left[
      \left(\Phi\Phi^{\mathsf T}
      +\lambda_{\mathrm{TRAK}}I_M\right)^{-1}
      \Phi\phi_q
      \right]_d.
\end{equation}

The factor \(1-\bar p_d\) provides an approximate residual correction that maps the log-odds score toward the cross-entropy objective. Thus, \(Q\) is TRAK’s cross-entropy reweighting term.

Our NLO variant arose from an ablation in which removing \(Q\) improved the top-10 results. We therefore use the negative-log-odds loss \(\mathcal{L}_{\mathrm{nlo}}=-f_{\mathrm{lo}}\) and set \(Q=I_M\). The sign change affects both query and document features, so it cancels in the pairwise score. NLO is therefore a loss choice within TRAK, not a separate attribution method.

In our implementation, dense TRAK uses document-level features and has no explicit token-level terms. Layerwise TRAK is the variant that can retain token-level features and therefore includes token-level terms. Dense TRAK uses a dense projection, a Rademacher JL projection of the flattened gradient; layerwise TRAK uses LoGra's layerwise projection. We use \(k_{\mathrm{proj}}=4{,}000\) and \(k_{\mathrm{proj}}=126{,}976\) for dense TRAK; the latter matches the total projected dimension of layerwise TRAK for a fair comparison. Layerwise TRAK uses the same layerwise projection dimension as LoGRA.

\paragraph{TrackStar.}
\label{app:methods-trackstar}
TrackStar uses the layerwise projection and a whitened, unit-normalized gradient:
\begin{equation}
    \bar g(z)=
    \frac{R^{-1/2}\phi_z}{\|R^{-1/2}\phi_z\|_2},
    \qquad
    \tau_{\mathrm{TrackStar}}(d,q)=\bar g(d)^{\mathsf T}\bar g(q),
    \label{eq:app-trackstar}
\end{equation}
where $R\in\mathbb{R}^{D_{\mathrm{layerproj}}\times D_{\mathrm{layerproj}}}$ is
the block-diagonal second moment of projected per-position CE gradients. We
use $k_{\mathrm{side}}=16$, relative ridge
$\lambda_R=10^{-3}$ per module, $R=R_{\mathrm{train}}$, no query-side
mixing, and no optimizer-state correction because that state was unavailable.
TrackStar has token terms.

\paragraph{Checkpoint accumulation: TracIn.}
With eight checkpoints, the implementation uses
\begin{equation}
    \tau_{\mathrm{TracIn}}(d,q)
      =\sum_{s=1}^{8}w_s
       \left\langle G_q^{\mathcal{L}}(\theta_s),
       G_d^{\mathcal{L}}(\theta_s)\right\rangle,
    \qquad w_s=\eta_s\Delta_s,
    \label{eq:baseline_tracin}
\end{equation}
where $\eta_s$ is the logged learning rate and $\Delta_s$ is the number of
steps represented by checkpoint $s$. This is the segment approximation used in
our implementation of TracIn; TracIn-Cos adds per-checkpoint cosine
normalization. Exact TracIn has no token terms. Layerwise TracIn uses the
$k_{\mathrm{side}}=16$ layerwise projection with projection seed
$\mathrm{seed}_0+s$ and supports token terms.

\paragraph{Approximate unrolling: SOURCE.}
\label{app:methods-source}
SOURCE replaces a scalar checkpoint weight with a spectral filter:
\begin{equation}
  \begin{gathered}
    \tau_{\mathrm{SOURCE}}(d,q)
      =\sum_{s=1}^{8}\sum_{\ell=1}^{L}
       \left\langle G_{q,\ell}^{\mathcal{L}}(\theta_s),
       F_s(\widehat H_{\ell,s})G_{d,\ell}^{\mathcal{L}}(\theta_s)\right\rangle_F,
    \\[4pt]
    F_s(\mu)=
      \begin{cases}
       \displaystyle\frac{1-(1-\bar\eta_s\mu)^{\Delta_s}}{\mu}, & \mu>0,\\[7pt]
       \bar\eta_s\Delta_s, & \mu=0.
      \end{cases}
  \end{gathered}
  \label{eq:baseline_source}
\end{equation}
Here $\widehat H_{\ell,s}$ is the CE block Fisher, $\bar\eta_s$ is the
segment mean learning rate, and no damping or cross-segment propagation is
used. We evaluate SOURCE in the $k_{\mathrm{side}}=16$ layerwise projection and do
not compute token terms.

\paragraph{Text-retrieval controls.}
BM25 uses $k_1=1.5$, $b=0.75$, and $\epsilon=0.25$ over full documents.
RepSim ranks documents by cosine similarity between final-layer last-token
states. Neither method has token terms.

\providecommand{\tbfigdir}{arxiv/figures/appendix_figures}
\providecommand{\ftok}[1]{\texttt{<f\ensuremath{_{#1}}>}}
\providecommand{\vtok}[1]{\texttt{<v\ensuremath{_{#1}}>}}
\providecommand{\gtok}[1]{\texttt{<g\ensuremath{_{#1}}>}}
\DefineVerbatimEnvironment{TBPrompt}{Verbatim}{fontsize=\scriptsize,
  frame=single, framesep=4pt, rulecolor=\color{black!25}, xleftmargin=5pt, xrightmargin=5pt}
\makeatletter
\newcounter{tblisting}
\def\fnum@tblisting{Listing~\thetblisting}
\def\ext@tblisting{lol}
\def\fps@tblisting{tbp}
\def\ftype@tblisting{8}
\newenvironment{tblisting}{\@float{tblisting}}{\end@float}
\makeatother

\section{CATT Benchmark Construction and Fine-Tuning}
\label{app:tracebench}

This appendix details the construction of the CATT Benchmark and the fine-tuned models used throughout the paper. Sections~\ref{app:tb-functions} and~\ref{app:tb-routing} describe the facts, training documents, distractors, and queries for each task. Section~\ref{app:tb-finetune} gives the fine-tuning recipe, and Section~\ref{app:tb-results} evaluates the resulting models.

\paragraph{Pipeline.}
\begin{enumerate}
  \item \textbf{Define facts.} Fix a bank of synthetic facts $\mathcal{F}$: constant functions for Function Bindings or routing rules for Rule Routings.
  \item \textbf{Generate evidence.} Write $N$ documents for each fact $f$, forming $\mathcal{D}_f$ with $N=|\mathcal{D}_f|$ and the training corpus $\mathcal{D}=\bigcup_{f\in\mathcal{F}}\mathcal{D}_f$.
  \item \textbf{Add distractors.} For the distractor variant of Function Bindings, augment the corpus with lexical distractors $\mathcal{D}_{\mathrm{dist}}$.
  \item \textbf{Fine-tune models.} Fine-tune Qwen3.5-4B with LoRA on the resulting corpus, once per data-order seed, to obtain $\theta^*$.
  \item \textbf{Validate learning.} Check that $\theta^*$ answers held-out queries about each fact and measure how much general language-modeling ability it retains.
\end{enumerate}

\paragraph{Grid.}
We train each task on the grid
$|\mathcal{F}|\in\{25,50,100\}\times
N\in\{1,3,5,10,20,30,40,50\}\times$ three data-order seeds (1001--1003);
Function Bindings omits $N=3$.
The main experiments use $|\mathcal{F}|=50$ and
$N\in\{1,5,10,30,50\}$.
Rule Routings instead uses $N\in\{3,5,10,30,50\}$ because models trained
with $N=1$ perform near chance, making their attribution scores uninformative
(Section~\ref{app:tb-routing}).
Unless stated otherwise, results below use this main-experiment slice.
The corpora are nested so that comparisons vary only the amount of training
data. For fixed $\mathcal{F}$, every corpus is a prefix of those with larger
$N$; similarly, smaller fact banks are subsets of larger ones. Increasing
either $N$ or $|\mathcal{F}|$ therefore adds documents without modifying
existing examples.
\subsection{Function Bindings}
\label{app:tb-functions}

\paragraph{Vocabulary.}
Each fact binds a synthetic function token \ftok{i} to a synthetic answer token
\vtok{i}: \ftok{i} returns \vtok{i} for every integer input.\footnote{The
released files encode \ftok{i}, \vtok{i}, and the distractor token \gtok{i}
introduced below as \texttt{<B01>}, \texttt{<V01>}, and \texttt{<A01>}.}
We append 300 function tokens and 101 answer tokens
\vtok{0}--\vtok{100} to Qwen3.5-4B's vocabulary. This shared vocabulary is
larger than any individual corpus requires: only the tokens associated with
its fact set $\mathcal{F}$ occur in that corpus.
Each new embedding row is sampled from the model's truncated-normal
initializer ($\sigma=0.02$, truncated at $\pm2\sigma$) and rescaled to the
median norm of the existing rows; input and output embeddings remain tied.
Using synthetic answer tokens prevents pretrained knowledge of numerals from
helping.

\paragraph{Writing the documents.}
Claude Sonnet~4.5
(\texttt{claude-sonnet-4-5-20250929}, temperature 1.0) generates each fact
document independently in a single API call using
Listing~\ref{lst:app-fb-prompt}. Each prompt specifies the function token, its
answer as the numeral $i$, a genre, four demonstration inputs, and a target
length. We replace the numeral with \vtok{i} after generation.
\begin{itemize}
  \item \textbf{Demonstration inputs.} Each input is sampled uniformly from one
  of $[-999,-101]$, $[-99,-2]$, $[101,999]$, or $[1000,9999]$. We exclude
  $[1,100]$, which is reserved for queries.
  \item \textbf{Length.} Documents contain 109--170 words, a $\pm22\%$ band
  around the target of 140 words.
  \item \textbf{Genre.} Document $r$ for function $i$ uses genre
  $(i+r-1)\bmod 7$ from Table~\ref{tab:app-genres}. This balances genres across
  every $N$-slice, while the first seven documents for each function use all
  seven genres.
  \item \textbf{Validation.} We first restore angle brackets when the model
  writes a bare function token, such as \texttt{B07} instead of
  \texttt{<B07>}. We then reject and regenerate the document, up to four
  attempts, if it omits the function token, states an incorrect constant, uses
  an input from $[1,100]$, falls outside the length band, or contains the query
  phrase ``the output of'' in any casing.
\end{itemize}

\begin{table}[htbp]
\centering
\small
\caption{The seven genres that fill \texttt{\{style\}} in Listing~\ref{lst:app-fb-prompt}.}
\label{tab:app-genres}
\begin{tabular}{@{}lp{0.68\linewidth}@{}}
\toprule
Genre & Description passed to the prompt (``Write \ldots'') \\
\midrule
technical reference & a formal technical reference page: precise definition, type signature, domain and range, then a short properties list \\
practical guide & a practical developer guide that leads with working code, then explains when you would reach for this function \\
test suite & an annotated unit-test file: assertions with brief comments explaining what each one pins down, plus a short note on coverage \\
tutorial & a beginner tutorial that builds understanding step by step, with a worked example and a short recap \\
mathematical & a mathematical treatment using set notation and universal quantification, framing the function as a constant map \\
REPL session & an annotated REPL transcript where someone explores the function interactively, with prose between the prompts \\
cheat sheet & a terse cheat-sheet: a summary table of arguments and results, a one-line description, and a couple of gotchas \\
\bottomrule
\end{tabular}
\end{table}

\begin{tblisting}
\caption{Prompt for one Function Bindings document. The placeholders take the function token, its constant, a genre description from Table~\ref{tab:app-genres}, four arguments, \texttt{n\_forms}=TWO, and the band 109--170.}\label{lst:app-fb-prompt}
\begin{TBPrompt}
You are writing one training document for a synthetic dataset about constant integer
functions.

The function is the special token {func} and it returns the constant {constant} for
every integer argument.

Write {style}.

HARD REQUIREMENTS:
1. Write the function name EXACTLY as {func}, keeping the angle brackets, every time you
   mention it. Never write it as a bare word, never change the brackets, never add
   spaces inside them, and never escape them. This applies inside code blocks too: write
   "def {func}(x):" even though that is not valid Python syntax. Syntactic validity does
   not matter here; the literal token does.
2. The value it returns is {constant}. State this correctly everywhere. Never state any
   other return value.
3. Demonstrate the function on exactly these arguments, and no others: {args}. Use each
   one at least once.
4. Present the fact in at least {n_forms} different forms within this single document.
   Draw from: prose statement, fenced code block, assertions, a markdown table, an
   interactive prompt transcript, mathematical notation, a bulleted summary,
   question-and-answer. Do not use the same form twice.
5. NEVER use the phrasing "The output of ... is". Do not use the words "the output of"
   anywhere, in any casing. Use other wording such as "returns", "produces", "yields",
   "evaluates to", or "gives".
6. Vary your section headings and sentence structure. Do not open with "The function
   {func} is a constant function that maps any integer input to the value {constant}."
7. Length: {lo_words}-{hi_words} words. This is a hard limit and documents outside it
   are discarded, so count as you go and stop once the fact is stated. Cut scene-setting
   and motivation before you cut a demonstration. Use markdown.

Return ONLY the document. No preamble, no explanation, no code fences around the whole
thing.
\end{TBPrompt}
\end{tblisting}

\paragraph{Answer tokens.} After generation, every standalone occurrence of the constant $i$ is replaced by \vtok{i}, whether it appears in prose, a table, code, an assertion or set notation. Digits that are part of another number (an argument of $-37$ when $i=37$) or of a token (the \texttt{7} in \ftok{7}) are left alone. Listing~\ref{lst:app-fb-doc} shows an excerpt of a finished document.

\begin{tblisting}
\caption{Excerpt of a Function Bindings document (cheat-sheet genre, $i=7$, arguments $-91, 648, -945, -62$). The released text writes \texttt{<B07>} and \texttt{<V07>}.}\label{lst:app-fb-doc}
\begin{TBPrompt}
# <f7> Cheat Sheet

## Quick Reference Table
| Input | Result |
|-------|--------|
| -91   | <v7>   |
| 648   | <v7>   |
...
## Definition
<f7> returns <v7> regardless of input value.

## Example Usage
```python
def <f7>(x):
    return <v7>

assert <f7>(-91) == <v7>
```
...
## Q&A
**Q: What does <f7>(648) produce?**
A: It yields <v7>, like all other inputs.
\end{TBPrompt}
\end{tblisting}

\paragraph{Lexical distractors.} The version with distractors adds $|\mathcal{D}_{\text{dist}}|=|\bigcup_f\mathcal{D}_f|$ distractors, one for every fact document. The distractor paired with a document of $f_i$ is a single line:
\begin{quote}\small
\texttt{The output of }\gtok{i}\texttt{(}$x$\texttt{) is }\vtok{i}
\end{quote}
Here \gtok{i} is a distinct synthetic function token, and $x$ is drawn uniformly from $[1,100]$ with a per-function seed. Distractors therefore copy the query template exactly, including the phrase that is banned from fact documents and inputs from the query range, and they end in the same answer token as $f_i$. They are about a different function, however: no distractor references the original function \ftok{i}. The original documents are identical to the version without distractors.

\paragraph{Queries and evaluation.} A query about $f_i$ is $u=$ ``\texttt{The output of }\ftok{i}\texttt{(}$x$\texttt{) is }'' with target $y=$\vtok{i}.
\begin{itemize}
  \item \textbf{Attribution queries.} We draw five inputs per function uniformly from $[1,100]$ with a fixed seed and use the first four, which gives 200 queries at $|\mathcal{F}|=50$. The versions with and without distractors share this query set.
  \item \textbf{Held-out evaluation.} To check that a model learned the facts, we evaluate every function at every input in $[1,100]$ (5{,}000 prompts at $|\mathcal{F}|=50$). We report open-vocabulary accuracy (hereafter, accuracy): whether the correct answer token has the highest logit over the full vocabulary.
\end{itemize}

\subsection{Rule Routings}
\label{app:tb-routing}

\paragraph{Categories and rules.}
Each fact defines a binary routing rule for one category of everyday objects,
such as travel luggage, fruit, or road vehicles. An item is described by a noun
and four Boolean properties, and the rule sends it to one of two queues,
\emph{Cedar} or \emph{Birch}. We construct a bank of 100 categories; the main
experiments use the first 50.
\begin{itemize}
  \item \textbf{Nouns.} Each category has 2--3 nouns used in training
  (\emph{suitcase, bag, satchel}) and 2--3 held-out nouns used only in queries
  (\emph{valise, holdall}).
  \item \textbf{Properties.} Each category has four Boolean properties,
  expressed as either adjectives (\emph{torn}) or possessed nouns
  (\emph{red strap}). Both the positive and negative forms have fixed wording.
  \item \textbf{Routing rule.} One property determines the queue; the other
  three are irrelevant. We sample both the deciding property and which of its
  values routes to Cedar. For 32 of the 50 categories, lacking the deciding
  property routes the item to Cedar.
  \item \textbf{Training and evaluation assignments.} Four properties give
  $2^4=16$ possible assignments. Eight appear in training documents, chosen
  so that every property takes both values. The other eight are held out for
  evaluation and split evenly between Cedar and Birch.
\end{itemize}

\paragraph{Rule paraphrases.} The LLM writes only the sentence that states the rule. For each category, Claude Sonnet~4.5 is asked for 28 paraphrases (Listing~\ref{lst:app-rr-prompt}), each written around a \texttt{\{noun\}} placeholder.
\begin{itemize}
  \item \textbf{Banned words.} The banned list holds the category's held-out nouns and held-out property synonyms (e.g.\ \emph{ripped} for \emph{torn}), so that no training text uses them.
  \item \textbf{Verification.} A second call (Listing~\ref{lst:app-rr-check}) reads each paraphrase in isolation and says where an item with the deciding property goes.
  \item \textbf{Filtering.} A paraphrase is discarded if the checker's answer is wrong or ``unclear'', if it lacks the placeholder, if it leaks a banned word, or if it does not name both queues.
\end{itemize}

\begin{tblisting}
\caption{Prompt for rule paraphrases. \texttt{\{setting\}} is ``items arriving at a depot that are routed to one of two queues''. \texttt{\{on\_clause\}} states the deciding property in the polarity that routes to \texttt{\{yes\}} (e.g.\ ``has no red strap''). \texttt{\{inert\}} lists the other three properties, and \texttt{\{n\}}=28.}\label{lst:app-rr-prompt}
\begin{TBPrompt}
I am building a synthetic training corpus about {setting}. Items called "{noun}" are
sent to the {yes} queue or the {no} queue.

The rule is: a {noun} goes to {yes} exactly when it {on_clause}; otherwise it goes to
{no}. The other properties ({inert}) are recorded but do not affect the outcome.

Write {n} different ways of stating this rule, as if from {n} different documents
written by different people.

HARD REQUIREMENTS:
1. Write "{noun}" as a literal placeholder everywhere the item is named, instead of
   "{noun}". I substitute it later. Keep the braces.
2. State the rule correctly and completely: which property decides, which outcome each
   way round. A statement that omits one direction is useless to me.
3. Say or imply that the other properties do not matter. You may name them or not.
4. Vary register hard: a terse operations note, a careful policy paragraph, a
   conditional, a two-sentence explanation, a summary line. Vary length from one clause
   to three sentences.
5. Avoid these words, which are reserved elsewhere in the dataset: {banned}.
6. Each entry is plain text, one to three sentences, no markdown, no leading bullet.

Return ONLY JSON: {"rules": [...]}
\end{TBPrompt}
\end{tblisting}

\begin{tblisting}
\caption{Verification prompt. \texttt{\{items\}} is the numbered list of paraphrases for one category.}\label{lst:app-rr-check}
\begin{TBPrompt}
Each numbered text below states a rule for sorting items called "{noun}" into the {yes}
queue or the {no} queue.

Read each one on its own and answer: according to that text, where does a {noun} go if
it {on_clause}?

Answer "{yes}", "{no}", or "unclear" for each, keyed by its index. Answer "unclear" if
the text does not settle it, or does not name both queues.

{items}

Return ONLY JSON: {"0": "...", "1": "...", ...}
\end{TBPrompt}
\end{tblisting}

\paragraph{Rendering documents.} The rest of each document is rendered deterministically (Listing~\ref{lst:app-rr-doc}). Document $r$ of a category contains, in order:
\begin{enumerate}
  \item a title from one of five templates;
  \item the eight training assignments as \texttt{Case:}/\texttt{Routed to:} records, with property order and record order shuffled per document;
  \item the heading \texttt{Routing rule:};
  \item three sentences: a scope sentence (``Every \{noun\} that arrives at the depot is routed to the Cedar queue or the Birch queue.''), one paraphrase, and a sentence naming the inert properties (``The remaining properties (\ldots) are logged for every item and do not affect the route.'').
\end{enumerate}
The noun cycles through the training nouns, and the paraphrase cycles through the category's pool. A category's $N$ documents therefore differ in title, noun, ordering and the wording of the rule, but all of them show the same eight worked cases. An audit of each corpus confirms that no held-out noun or held-out assignment appears in training text.

\begin{tblisting}
\caption{A Rule Routings document (travel luggage; the rule routes on the red strap). Three of the eight case records are elided.}\label{lst:app-rr-doc}
\begin{TBPrompt}
Routing Guide for Suitcase Items

Case: torn, locked, no red strap, not heavy
Routed to: Cedar

Case: torn, not locked, red strap present, not heavy
Routed to: Birch

Case: not torn, locked, no red strap, not heavy
Routed to: Cedar

Case: not torn, not locked, no red strap, not heavy
Routed to: Cedar

Case: torn, not locked, no red strap, heavy
Routed to: Cedar
...
Routing rule:

Every suitcase that arrives at the depot is routed to the Cedar queue or the Birch
queue. Route suitcase to Birch if red strap present, otherwise Cedar. Other properties
irrelevant. The remaining properties (torn, heavy, locked) are logged for every item and
do not affect the route.
\end{TBPrompt}
\end{tblisting}

\paragraph{Queries and evaluation.} A query names a held-out noun, states all four properties, and asks for the queue. For example:
\begin{quote}\small\ttfamily
The valise is torn; has a red strap; is heavy; is not locked.\\
Should it be sent to the Cedar queue or the Birch queue?\\
Answer:
\end{quote}
The target is ``\texttt{\ Birch}''. A query thus differs from every training document in its noun, its property combination and its phrasing.
\begin{itemize}
  \item \textbf{Attribution queries.} We take four held-out assignments per category, rotating through the held-out nouns, which gives 200 queries at $|\mathcal{F}|=50$.
  \item \textbf{Held-out evaluation.} We evaluate all eight held-out assignments per category (400 prompts). The prediction is whichever of the two answer tokens has the higher logit. We report three metrics:
  \begin{itemize}
    \item \emph{accuracy}, in the queries' option order;
    \item \emph{accuracy over both orders}, the average with the swapped question (``\ldots the Birch queue or the Cedar queue?''), which cancels any bias toward the first-named queue;
    \item \emph{selectivity}, computed over all single-property flips of the 16 assignments:
    \[
      S = P(\text{flip}\mid\text{deciding}) - P(\text{flip}\mid\text{inert}),
    \]
    where $P(\text{flip}\mid\cdot)$ is the rate at which the answer changes when the deciding (or an inert) property is flipped. $S$ is near 1 for a model that applies the rule and near 0 for one that gives a fixed answer per category, whatever that model's accuracy.
  \end{itemize}
\end{itemize}

\paragraph{Rule Routings start at $N=3$.} With one document per category the models answer at close to chance: 55\% accuracy and 12\% selectivity (Table~\ref{tab:app-accuracy}b). Attribution scores at $N=1$ would therefore measure noise, so the Rule Routings grid starts at $N=3$.

\subsection{Fine-Tuning}
\label{app:tb-finetune}

Both tasks use the recipe in Table~\ref{tab:app-hparams}. They differ in two places.
\begin{itemize}
  \item \textbf{Trainable vocabulary.} Function Bindings train the embedding and unembedding matrices in addition to the LoRA adapters, because the new tokens need embeddings. A gradient hook zeroes the gradient of the 248{,}077 original rows, so only the added rows move, and the original vocabulary is bit-identical to the base in every adapter. Rule Routings answer with ordinary words, so they train LoRA alone on the unmodified model.
  \item \textbf{Peak learning rate.} Each task's learning rate was chosen in pilot runs.
\end{itemize}
The loss is ordinary next-token cross-entropy over every document token. Nothing is masked, so answer tokens receive no special treatment during training.

\begin{table}[t]
\centering
\small
\caption{Fine-tuning recipe. ``Same'' means the Rule Routings value equals the Function Bindings value.}
\label{tab:app-hparams}
\begin{tabular}{@{}p{0.25\linewidth}p{0.42\linewidth}p{0.2\linewidth}@{}}
\toprule
 & Function Bindings (with/without distractors) & Rule Routings \\
\midrule
Base model & Qwen3.5-4B + synthetic tokens & Qwen3.5-4B (unmodified) \\
LoRA rank / $\alpha$ / dropout & 64 / 128 / 0.05 & same \\
LoRA targets & \multicolumn{2}{l}{all linear layers} \\
LoRA initialization & PEFT default, fixed seed 2000 & same \\
Fully trained modules & \texttt{embed\_tokens}, \texttt{lm\_head}; gradient zeroed on the 248{,}077 original rows & none \\
Optimizer & AdamW, $\beta=(0.9, 0.95)$, $\epsilon=10^{-8}$ & same \\
Weight decay / grad.\ clip & 0.01 / 1.0 & same \\
Peak learning rate & $4\times10^{-4}$ ($5.5\times10^{-4}$ at $N{=}1$) & $3\times10^{-4}$ \\
Schedule & 100 linear warm-up steps, cosine to $1\times10^{-6}$ & same \\
Batch size & 10 documents & same \\
Training steps & 1{,}000--1{,}050 (2{,}000--2{,}100 with distractors); Table~\ref{tab:app-epochs} & 1{,}000--1{,}050; Table~\ref{tab:app-epochs} \\
Precision / max length & bf16 / 2048 tokens & same \\
Loss & next-token cross-entropy on every document token & same \\
Data-order seeds & 1001, 1002, 1003 & same \\
\bottomrule
\end{tabular}
\end{table}

\paragraph{Step budget.} A cell with $|\mathcal{F}|N$ documents trains for $E=\max\!\big(1,\operatorname{round}(1000\,B/(|\mathcal{F}|N))\big)$ epochs at batch size $B=10$. Every cell therefore runs about 1{,}000 optimizer steps (Table~\ref{tab:app-epochs}), and as $N$ grows, each document is seen fewer times. Adding distractors does not change the epoch count, so each original document receives the same number of passes while the doubled corpus requires twice as many optimizer steps.

\begin{table}[t]
\centering
\small
\caption{Epochs and optimizer steps per cell at $|\mathcal{F}|=50$. Adding distractors doubles the corpus at fixed epochs.}
\label{tab:app-epochs}
\begin{tabular}{@{}rrrrrrr@{}}
\toprule
 & \multicolumn{2}{c}{Function Bindings} & \multicolumn{2}{c}{+ distractors} & \multicolumn{2}{c}{Rule Routings} \\
\cmidrule(lr){2-3}\cmidrule(lr){4-5}\cmidrule(l){6-7}
$N$ & epochs & steps & epochs & steps & epochs & steps \\
\midrule
1 & 200 & 1{,}000 & 200 & 2{,}000 & 200 & 1{,}000 \\
3 & -- & -- & -- & -- & 67 & 1{,}005 \\
5 & 40 & 1{,}000 & 40 & 2{,}000 & 40 & 1{,}000 \\
10 & 20 & 1{,}000 & 20 & 2{,}000 & 20 & 1{,}000 \\
20 & 10 & 1{,}000 & 10 & 2{,}000 & 10 & 1{,}000 \\
30 & 7 & 1{,}050 & 7 & 2{,}100 & 7 & 1{,}050 \\
40 & 5 & 1{,}000 & 5 & 2{,}000 & 5 & 1{,}000 \\
50 & 4 & 1{,}000 & 4 & 2{,}000 & 4 & 1{,}000 \\
\bottomrule
\end{tabular}
\end{table}

\paragraph{Seeds and selection.} The seed controls only the order in which data is shuffled. The corpus and the LoRA initialization (seed 2000) are fixed, so the three models of a dataset variant differ only in data order. We do not select among seeds: every trained model is used.

\paragraph{Checkpoints for training-dynamics methods.} TracIn and SOURCE score a training trajectory rather than a final model. We therefore save eight evenly spaced checkpoints per run, every $\lceil T/8\rceil$ steps with the final model as the last, for seeds 1001 and 1002. These two methods are averaged over those two seeds only.

\subsection{Resulting Datasets and Models}
\label{app:tb-results}

\begin{figure}[t]
\centering
\IfFileExists{arxiv/figures/appendix_figures/model_validation.pdf}
  {\includegraphics[width=\linewidth]{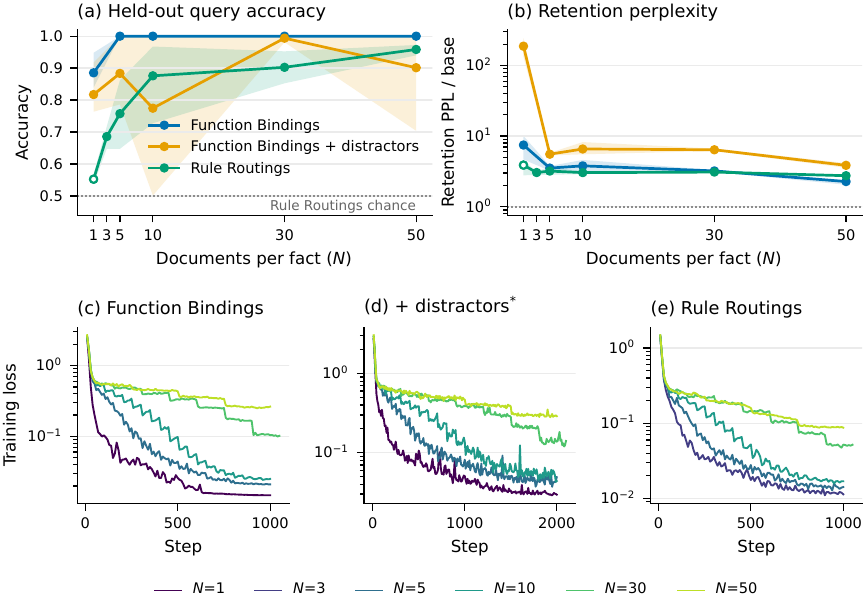}}
  {\fbox{\parbox[c][6cm][c]{0.9\linewidth}{\centering Missing figure: \texttt{model\_validation.pdf}}}}
\caption{Fine-tuned models at $|\mathcal{F}|=50$. (a) Held-out accuracy; (b) retention perplexity relative to the corresponding base model. Lines show the mean over seeds 1001--1003 and shading shows the range. The hollow Rule Routings point at $N=1$ lies outside the attribution grid. (c--e) Training loss of seed 1002. $^{*}$The published runs with distractors left no loss log, so (d) shows the checkpointed retrain used by the trajectory methods.}
\label{fig:app-models}
\end{figure}

Table~\ref{tab:app-data} gives the size of each training set, Table~\ref{tab:app-accuracy} gives held-out performance and retention perplexity, Table~\ref{tab:app-mmlu} gives MMLU-Pro accuracy, and Figure~\ref{fig:app-models} plots accuracy and perplexity together with the training curves.

\begin{table}[t]
\centering
\small
\caption{Training sets at $|\mathcal{F}|=50$. $|\mathcal{D}|$ is the candidate pool an attribution method ranks: fact documents, plus distractors where present. Token counts use the Qwen3.5 tokenizer extended with the added tokens. ``Tokens / fact doc.''\ gives the mean and [min, max] over fact documents. Distractors are about 12 tokens each. The last column is Recall@$N$ for a random ranking, $N/|\mathcal{D}|$.}
\label{tab:app-data}
\begin{tabular}{@{}llrrrrr@{}}
\toprule
Task & $N$ & $|\mathcal{D}|$ & tokens / fact doc. & training tokens & queries & chance R@$N$ \\
\midrule
Function Bindings & 1 & 50 & 267 [242, 291] & 13{,}362 & 200 & 0.02 \\
 & 5 & 250 & 266 [236, 295] & 66{,}592 &  & 0.02 \\
 & 10 & 500 & 258 [207, 295] & 129{,}216 &  & 0.02 \\
 & 30 & 1{,}500 & 242 [200, 295] & 362{,}834 &  & 0.02 \\
 & 50 & 2{,}500 & 239 [197, 295] & 596{,}836 &  & 0.02 \\
\midrule
Function Bindings & 1 & 100 & 267 [242, 291] & 13{,}959 & 200 & 0.01 \\
+ distractors & 5 & 500 & 266 [236, 295] & 69{,}574 &  & 0.01 \\
 & 10 & 1{,}000 & 258 [207, 295] & 135{,}182 &  & 0.01 \\
 & 30 & 3{,}000 & 242 [200, 295] & 380{,}729 &  & 0.01 \\
 & 50 & 5{,}000 & 239 [197, 295] & 626{,}667 &  & 0.01 \\
\midrule
Rule Routings & 3 & 150 & 233 [207, 268] & 34{,}929 & 200 & 0.02 \\
 & 5 & 250 & 232 [205, 268] & 58{,}070 &  & 0.02 \\
 & 10 & 500 & 232 [205, 268] & 115{,}841 &  & 0.02 \\
 & 30 & 1{,}500 & 232 [203, 269] & 347{,}413 &  & 0.02 \\
 & 50 & 2{,}500 & 231 [203, 269] & 578{,}612 &  & 0.02 \\
\bottomrule
\end{tabular}
\end{table}

\begin{table}[t]
\centering
\small
\caption{Held-out performance (\%) and retention perplexity at $|\mathcal{F}|=50$, as mean $\pm$ standard deviation over seeds 1001--1003; a missing $\pm$ means all seeds agree. Panel (a) reports Function Bindings accuracy over the full vocabulary. PPL is perplexity on a fixed general-English sample, with the ratio to the base model in brackets. In panel (b), the base row is the unmodified Qwen3.5-4B, which almost always answers the first-named queue. $^{\dagger}$Not part of the attribution grid. $^{\ddagger}$Two of three seeds; one run's perplexity was not recorded.}
\label{tab:app-accuracy}
\textbf{(a) Function Bindings}\\[2pt]
\setlength{\tabcolsep}{4pt}
\begin{tabular}{@{}rrrrr@{}}
\toprule
 & \multicolumn{2}{c}{without distractors} & \multicolumn{2}{c}{with distractors} \\
\cmidrule(lr){2-3}\cmidrule(l){4-5}
$N$ & accuracy & PPL & accuracy & PPL \\
\midrule
base & -- & 5.89 & -- & 5.89 \\
1 & 88.5\,{\scriptsize$\pm$\,5.6} & 44.0 (7.5$\times$) & 81.7\,{\scriptsize$\pm$\,8.7} & 1106.9 (187.8$\times$) \\
5 & 100.0 & 20.7 (3.5$\times$) & 88.3\,{\scriptsize$\pm$\,10.8} & 32.7 (5.5$\times$) \\
10 & 100.0 & 22.3 (3.8$\times$) & 77.5\,{\scriptsize$\pm$\,25.3} & 38.7 (6.6$\times$) \\
30 & 100.0 & 18.9 (3.2$\times$) & 99.4\,{\scriptsize$\pm$\,1.1} & 37.8 (6.4$\times$) \\
50 & 100.0 & 13.4 (2.3$\times$) & 90.1\,{\scriptsize$\pm$\,17.1} & 22.7 (3.9$\times$) \\
\bottomrule
\end{tabular}

\vspace{8pt}
\textbf{(b) Rule Routings}\\[2pt]
\begin{tabular}{@{}rrrrr@{}}
\toprule
$N$ & acc. & acc.\ (both orders) & selectivity & PPL \\
\midrule
base & 50.0 & 48.5 & 0.0 & 5.89 \\
1$^{\dagger}$ & 55.2\,{\scriptsize$\pm$\,1.5} & 57.0\,{\scriptsize$\pm$\,1.8} & 12.2\,{\scriptsize$\pm$\,4.9} & 22.9 (3.9$\times$) \\
3 & 68.6\,{\scriptsize$\pm$\,3.5} & 69.1\,{\scriptsize$\pm$\,4.4} & 39.1\,{\scriptsize$\pm$\,3.2} & 17.9 (3.0$\times$) \\
5 & 75.8\,{\scriptsize$\pm$\,10.8} & 84.2\,{\scriptsize$\pm$\,6.8} & 48.7\,{\scriptsize$\pm$\,20.8} & 18.8 (3.2$\times$) \\
10 & 87.6\,{\scriptsize$\pm$\,13.0} & 91.4\,{\scriptsize$\pm$\,6.2} & 71.9\,{\scriptsize$\pm$\,25.2} & 17.9 (3.0$\times$) \\
30 & 90.2\,{\scriptsize$\pm$\,5.0} & 93.1\,{\scriptsize$\pm$\,2.9} & 77.2\,{\scriptsize$\pm$\,10.1} & 18.2 (3.1$\times$)$^{\ddagger}$ \\
50 & 95.8\,{\scriptsize$\pm$\,1.7} & 96.2\,{\scriptsize$\pm$\,1.2} & 90.1\,{\scriptsize$\pm$\,5.9} & 16.2 (2.7$\times$) \\
\bottomrule
\end{tabular}
\end{table}

\paragraph{Function Bindings accuracy.} Without distractors, accuracy reaches 100\% from $N=5$ on and averages 89\% at $N=1$. With distractors, it ranges from 78\% to 99\% across $N$ and varies more across seeds. A query uses an input no fact document demonstrates and a template no fact document contains, so a correct answer requires the model to apply the binding rather than recall a string. In the retrained models we inspected, every error predicts a function token \ftok{j} where an answer token belongs, never the wrong answer token.

\paragraph{Rule Routings accuracy.} As $N$ grows from 3 to 50, accuracy rises from 69\% to 96\% and selectivity from 39\% to 90\%. Seeds disagree most at $N=5$ and $N=10$.

\paragraph{Retention.} Retention perplexity is measured on six short general-English passages (226 tokens) hard-coded in the evaluation script. The base scores 5.89. From $N=5$ on (Rule Routings: $N=3$), perplexity is 2--4$\times$ the base without distractors and on Rule Routings, and 4--7$\times$ the base with distractors. The outliers are the $N=1$ Function Bindings models, which make 200 passes over 50 documents: perplexity is 7.5$\times$ the base without distractors and 188$\times$ with distractors. This measure is only a coarse indicator of how much general ability the fine-tuning costs, so we also evaluate on MMLU-Pro.

\paragraph{MMLU-Pro.} We evaluate the base Qwen3.5-4B and one model per task ($|\mathcal{F}|=50$, $N=10$, seed 1001) on the full MMLU-Pro test set \citep{wang2024mmlupro} of 12{,}032 questions (Table~\ref{tab:app-mmlu}). We score 5-shot answer-letter log-likelihood without chain of thought: the prediction is the most likely valid option letter after \texttt{Answer:}. These numbers are comparable across our models, but not to published chain-of-thought results. Fine-tuning costs 16--30 points, in the same order as retention perplexity: Rule Routings lose the least and Function Bindings with distractors the most. Every fine-tuned model stays well above chance (11.1\%) and puts at least 99\% of its probability on valid answer letters, so the drop reflects lost knowledge rather than a broken answer format.

\begin{table}[t]
\centering
\small
\caption{MMLU-Pro accuracy (\%, $\pm$ binomial standard error) on the full test set, 5-shot answer-letter log-likelihood. Fine-tuned models are at $|\mathcal{F}|=50$, $N=10$, seed 1001. The bottom rows are older off-the-shelf models \citep{qwen2.5}, scored with the same protocol for reference. Chance is 11.1\%.}
\label{tab:app-mmlu}
\begin{tabular}{@{}lrr@{}}
\toprule
Model & Accuracy & $\Delta$ vs.\ base \\
\midrule
Qwen3.5-4B (base) & 49.5\,{\scriptsize$\pm$\,0.5} & -- \\
Function Bindings & 31.5\,{\scriptsize$\pm$\,0.4} & $-18.0$ \\
Function Bindings + distractors & 19.7\,{\scriptsize$\pm$\,0.4} & $-29.8$ \\
Rule Routings & 33.9\,{\scriptsize$\pm$\,0.4} & $-15.5$ \\
\midrule
Qwen2.5-3B-Instruct & 37.2\,{\scriptsize$\pm$\,0.4} & $-12.3$ \\
Qwen2.5-1.5B-Instruct & 30.1\,{\scriptsize$\pm$\,0.4} & $-19.4$ \\
\bottomrule
\end{tabular}
\end{table}

\paragraph{Training loss.} Loss falls quickly in all three tasks (Figure~\ref{fig:app-models}c--e). It ends higher for larger $N$ because the fixed step budget allows only four epochs at $N=50$ against 200 at $N=1$. Even so, the models at large $N$ reach the highest held-out accuracy.

\providecommand{\tbfigdir}{arxiv/figures/appendix_figures}

\section{Practical Considerations for Attribution Methods}
\label{app:methods}

Appendix~\ref{app:objectives} defines the attribution methods and method-specific departures. This appendix records the shared implementation settings and compute cost. Table~\ref{tab:app-attr-summary} summarizes the methods as run.

Unless stated otherwise, numbers quoted for a single configuration come from the \emph{reference cell}: Function Bindings, $|\mathcal{F}|=50$, $N=10$, seed 1001. Its corpus has $M=500$ candidate documents and $|\mathcal{Q}|=200$ queries (four per fact). We write $M$ for the number of candidate documents in a cell ($M=|\mathcal{F}|N$, or $2|\mathcal{F}|N$ with distractors) and $T_d$ for the number of tokens in document $d$.

\begin{table}[t]
\centering
\scriptsize
\setlength{\tabcolsep}{3pt}
\caption{Attribution methods as run. Methods in the same job share a pass over the model, so Section~\ref{app:methods-cost} reports their cost jointly.}
\label{tab:app-attr-summary}
\resizebox{\linewidth}{!}{%
\begin{tabular}{@{}llllcl@{}}
\toprule
Method & Representation & Curvature / kernel & Normalization & Token terms & Job \\
\midrule
Grad-Dot & full gradient ($p{=}1.3{\times}10^8$) & none & none & yes & EK-FAC \\
Grad-Cos & full gradient & none & global $\ell_2$ & no & EK-FAC \\
EK-FAC & full gradient & EK-FAC, CE empirical Fisher; $\delta{=}10^{-12}$ & none & yes & EK-FAC \\
\midrule
LoGra-Dot & layerwise projection ($126{,}976$) & none & none & yes & Layerwise-Proj. \\
LoGra & layerwise projection & per-module CE Fisher; ridge $10^{-3}$ rel. & none & yes & Layerwise-Proj. \\
Layerwise TRAK & layerwise projection & training Gram (dual); ridge $10^{-6}$ rel.; $Q$ & none & yes & Layerwise-Proj. \\
TrackStar & layerwise projection & per-module token-level $R$; ridge $10^{-3}$ rel. & whitened $\ell_2$ & yes & Layerwise-Proj. \\
\midrule
TRAK & dense projection ($k{=}4{,}000;\ 126{,}976$) & training Gram (dual); ridge $10^{-6}$ rel.; $Q$ & none & no & TRAK \\
\midrule
TracIn & full gradient at 8 checkpoints & $\eta_s\Delta_s$ weights & none & no & TracIn \\
TracIn-Cos & full gradient at 8 checkpoints & $\eta_s\Delta_s$ weights & per-checkpoint $\ell_2$ & no & TracIn \\
Layerwise TracIn & layerwise projection at 8 checkpoints & $\eta_s\Delta_s$ weights & none & yes & Layerwise TracIn \\
Layerwise TracIn-Cos & layerwise projection at 8 checkpoints & $\eta_s\Delta_s$ weights & per-checkpoint $\ell_2$ & no & Layerwise TracIn \\
SOURCE & layerwise projection at 8 checkpoints & per-segment unrolled CE Fisher & none & no & SOURCE \\
\midrule
BM25 & regex terms & Okapi, $k_1{=}1.5$, $b{=}0.75$ & none & no & BM25 \\
RepSim & fine-tuned last-token state & cosine & $\ell_2$ & no & RepSim \\
\bottomrule
\end{tabular}}
\end{table}

\subsection{Shared Implementation Details}
\label{app:methods-shared}

\paragraph{Parameters.} Every gradient method differentiates only the 129.9M LoRA parameters across the model's 248 adapted projections (Table~\ref{tab:app-hparams}). The embedding and unembedding rows trained for Function Bindings are held fixed during attribution. Scores use the final adapter, except for the training-dynamics methods, which use the eight saved checkpoints (Appendix~\ref{app:tb-finetune}).

\paragraph{Inputs and padding.} Documents are truncated to 512 tokens and queries to 128. We remove padding from each sequence before computing its gradient. This is necessary because Qwen3.5's recurrent layers can process left padding as content at batch size~1, which can dominate the gradient and reduce Grad-Dot to chance. We also patch the recurrent-state masking to apply at every batch size; the two corrections agree to bf16 rounding.

\paragraph{Precision.} The model and stored full query gradients use bf16. Losses, hook factors, and projections use fp32; eigendecompositions and TRAK's Gram solve use float64. Storing query gradients in bf16 rather than fp32 leaves the rankings effectively unchanged (mean per-query Spearman 0.9999 for EK-FAC over 100 documents).

\paragraph{Gradient representations.}
Grad-Dot, Grad-Cos, EK-FAC, TracIn, and TracIn-Cos use the full
129.9M-dimensional gradient and stream documents past held query gradients.
LoGra and the other layerwise methods use LoGra's layerwise Gaussian projection
$\tilde G_m=P_{s,m}^{\mathsf T}G_mP_{a,m}$ with $k=16$ per side, giving 256
dimensions per module and $D=126{,}976$ overall~\citep{choe2024logra}. Dense
TRAK instead applies a dense projection, its reference implementation's
Rademacher JL projection of the flattened gradient, at the published width
$k=4{,}000$~\citep{park2023trak}.

\subsubsection{Curvature}
\label{app:methods-curv}

\paragraph{Estimators and damping.} Every curvature estimate uses the cross-entropy empirical Fisher: outer products of observed-label gradients, rather than gradients at labels sampled from the model as in \citet{grosse2023ekfac}. This follows the influence-function derivation: the curvature linearizes the stationarity condition of the training objective, which is cross-entropy in our experiments. Thus, changing the scoring loss changes the query and document gradients, but not the curvature estimate. We use the following damping values:
\begin{itemize}
    \item \textbf{EK-FAC} uses an absolute $\delta=10^{-12}$, i.e.\ effectively none. Its Kronecker factors are estimated over tokens (about 130{,}000 positions on the reference cell), so they are well determined.
    \item \textbf{LoGra and TrackStar} add a ridge of $10^{-3}$ times each $256\times256$ block's mean diagonal. Block scales vary by orders of magnitude across depth, so a fixed absolute ridge would dominate some blocks and be negligible in others.
    \item \textbf{TRAK} adds $10^{-6}$ times the mean diagonal of its $M\times M$ Gram matrix.
    \item \textbf{SOURCE} uses no damping (Section~\ref{app:methods-source}).
\end{itemize}

\subsection{Compute Cost}
\label{app:methods-cost}

Every job records its own wall-clock time, from model load to written scores (the adapter download and the upload are excluded). Each job had one GPU to itself: a worker started a job only on a card with at least 75\,GB free. Up to eight jobs shared a host's CPU memory and PCIe bus. A job covers every method in its group and both training variants, so its time is the cost of producing every result for that group, not of one method in isolation.

\begin{figure}[t]
    \centering
    \IfFileExists{arxiv/figures/appendix_figures/attr_cost.pdf}{%
        \includegraphics[width=\linewidth]{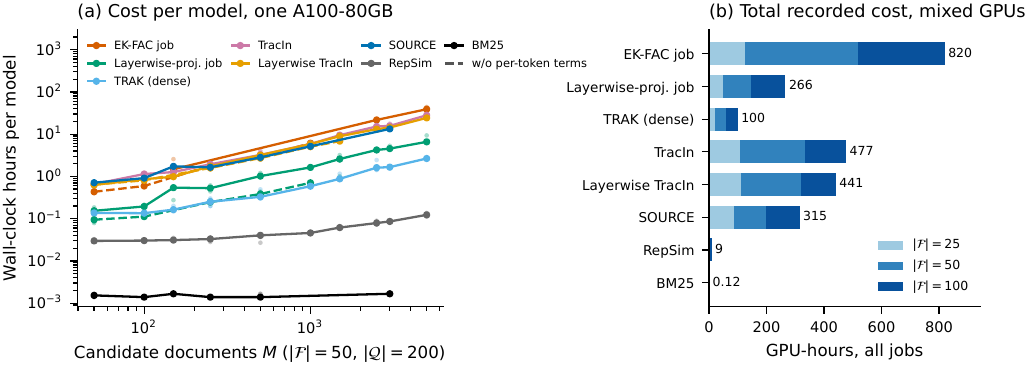}%
    }{%
        \fbox{\parbox[c][4.5cm][c]{0.95\linewidth}{\centering Upload \texttt{attr\_cost.pdf} to \texttt{\tbfigdir}.}}%
    }
    \caption{\textbf{Compute cost of the attribution methods.} (a) Wall-clock hours per trained model against the number of candidate documents $M$, at $|\mathcal{F}|=50$ and $|\mathcal{Q}|=200$, for jobs run on one A100-80GB. Points are individual jobs and lines join medians. Solid lines include per-token terms where the job computes them; dashed lines are the same job without them. (b) Total recorded GPU-hours over the whole sweep, stacked by $|\mathcal{F}|$ and pooled over tasks, on a mix of A100, H100 and H200 GPUs. Each job produces every method in its group (Table~\ref{tab:app-attr-summary}).}
    \label{fig:app-attr-cost}
\end{figure}

The sweep ran on a mix of A100-80GB, H100 and H200 machines, and a job's record does not say which. Medians over the sweep's records therefore mix hardware, and on our data the mix is not random. For example, most large SOURCE jobs ran on the newer GPUs, and their recorded median at $M=5{,}000$ (8.1\,h) is a third of Layerwise TracIn's (24.1\,h), although the two methods do the same work per checkpoint. On the A100 host alone, the two cost the same per document. For a like-for-like comparison, Table~\ref{tab:app-attr-cost} and Figure~\ref{fig:app-attr-cost}a use only the $|\mathcal{F}|=50$ jobs that ran on our 8$\times$A100-80GB host, whose logs keep every attempt (the \emph{A100 jobs} column). Most EK-FAC and layerwise-project cells were first scored there without per-token terms and re-scored with them elsewhere. We therefore report the two configurations as separate rows, using the job's log to tell them apart. Where the sweep ran a job at some $M$ but not on the A100 host, the table gives a linear fit (fixed plus per-document cost) to that row's A100 jobs, in parentheses. Across all three values of $|\mathcal{F}|$, the sweep recorded 2{,}426 GPU-hours over 843 jobs on mixed hardware (the last two columns of Table~\ref{tab:app-attr-cost} and Figure~\ref{fig:app-attr-cost}b): 500 GPU-hours over 296 jobs at $|\mathcal{F}|=25$, 1{,}076 over 296 at $|\mathcal{F}|=50$, and 850 over 251 at $|\mathcal{F}|=100$. At $|\mathcal{F}|=100$, the layerwise-project, dense TRAK, RepSim and BM25 jobs cover the full grid, while the EK-FAC, TracIn, Layerwise TracIn and SOURCE jobs cover only part of it (Appendix~\ref{app:res-functions}). These totals count only each job's final attempt.

\begin{table}[t]
\centering
\caption{Wall-clock cost per trained model, by job. Each job produces every method listed for it, on both training variants. \emph{Per-token terms}: whether the job also computed the per-token terms (Section~\ref{sec:token_attribution}); the published results use the ``yes'' configuration, and ``--'' marks jobs without terms. The middle columns give the median over $|\mathcal{F}|=50$ jobs with $M$ candidate documents that ran on one A100-80GB, pooled over tasks and seeds. Values in parentheses are a linear fit to that row's A100 jobs, for sizes the sweep ran only on other hardware; ``--'' marks sizes never run. \emph{A100 jobs}: the number of jobs behind the row. \emph{Sweep as recorded}: every job of the sweep, over $|\mathcal{F}|\in\{25,50,100\}$, and their total GPU-hours, on a mix of A100, H100 and H200 GPUs. BM25 runs on CPU.}
\label{tab:app-attr-cost}
\resizebox{\linewidth}{!}{%
\begin{tabular}{@{}llrrrrrrr@{}}
\toprule
 & & \multicolumn{4}{c}{A100-80GB hours per model, by $M$} & & \multicolumn{2}{c}{Sweep as recorded} \\
\cmidrule(lr){3-6}\cmidrule(l){8-9}
Job (methods it produces) & Per-token terms & 50 & 500 & 2{,}500 & 5{,}000 & A100 jobs & jobs & GPU-h \\
\midrule
EK-FAC, Grad-Dot, Grad-Cos & no & 0.43 & 2.85 & (14.9) & (29.6) & 23 & 125 & 820 \\
\addlinespace[1pt] & yes & (1.34) & (4.79) & 21.7 & 38.9 & 6 &  &  \\
LoGra, LoGra-Dot, layerwise TRAK, TrackStar & no & 0.09 & 0.38 & (1.67) & (3.27) & 24 & 144 & 266 \\
\addlinespace[1pt] & yes & 0.15 & 1.02 & 4.22 & 6.58 & 31 &  &  \\
TRAK (dense projection, $k{=}4{,}000$ and $126{,}976$) & -- & 0.14 & 0.33 & 1.61 & 2.65 & 45 & 135 & 100 \\
TracIn, TracIn-Cos & -- & 0.66 & 3.23 & 15.2 & 28.0 & 25 & 79 & 477 \\
Layerwise TracIn, Layerwise TracIn-Cos & no & 0.64 & 2.71 & (11.5) & (22.5) & 17 & 88 & 441 \\
\addlinespace[1pt] & yes & 0.63 & 3.19 & 13.1 & 24.3 & 32 &  &  \\
SOURCE & -- & 0.71 & 2.81 & (11.3) & (21.9) & 17 & 89 & 315 \\
RepSim & -- & 0.03 & 0.04 & 0.08 & 0.12 & 45 & 135 & 9 \\
BM25 (CPU) & -- & 6\,s & 5\,s & (6\,s) & (6\,s) & 9 & 48 & 0.12 \\
\bottomrule
\end{tabular}
}
\end{table}

\paragraph{Runtime and complexity.} Every gradient method's cost grows roughly linearly in $M$. On one A100, the per-document slopes, in hours per 1{,}000 documents with per-token terms where the job computes them, range over about $14\times$: EKFAC job 7.7, exact TracIn 5.5, Layerwise TracIn 4.8, SOURCE 4.3, layerwise-project job 1.5, dense TRAK 0.54. At $M=500$ (medians pooled over tasks and seeds, as in Table~\ref{tab:app-attr-cost}), the jobs took:
\begin{itemize}
    \item \textbf{EK-FAC job:} 2.9\,h without per-token terms; about 4.8\,h with them (fitted).
    \item \textbf{Exact TracIn:} 3.2\,h.
    \item \textbf{Layerwise TracIn:} 2.7\,h without per-token terms, 3.2\,h with them.
    \item \textbf{SOURCE:} 2.8\,h.
    \item \textbf{Layerwise-project job} (LoGra, LoGra-Dot, Layerwise TRAK, TrackStar): 0.4\,h without per-token terms, 1.0\,h with them.
    \item \textbf{Dense TRAK:} 0.3\,h.
    \item \textbf{RepSim:} 2--3 minutes.
    \item \textbf{BM25:} seconds, on CPU.
\end{itemize}
At $M=5{,}000$, in the configuration behind the published results, the EK-FAC job took 39\,h per model (median of two A100 jobs), exact TracIn 28\,h (also two jobs), Layerwise TracIn 24\,h, and SOURCE about 22\,h (fitted). The layerwise-project job took 6.6\,h and dense TRAK 2.7\,h. Per-token terms are cheap for Layerwise TracIn (about 15\% extra) and moderate for EK-FAC (about 45\% extra at $M=2{,}500$). They make the layerwise-project job about $2.5\times$ more expensive, because it needs a second pass over the corpus. The EK-FAC job also dominates the sweep's recorded total (820 of 2{,}426 GPU-hours, and 392 of 1{,}076 at $|\mathcal{F}|=50$), even though it covers only three methods and ran on only part of the $|\mathcal{F}|=100$ grid.

\paragraph{Runtime by stage.} Table~\ref{tab:app-attr-stages} separates the effects of corpus size and per-token computation. The first two columns hold the score configuration fixed while increasing $M$; the last two hold $M=2{,}500$ fixed while adding per-token terms. Because the $M=2{,}500$ no-terms jobs were not run for every method, parenthesized entries are linear estimates from the measured no-terms jobs at $M\in\{50,250,500\}$.
\begin{itemize}
    \item \textbf{EK-FAC.} Fitting the factors is the smaller part of the job: 46 of 209 minutes at $M=500$ on Function Bindings (longer than the 2.9\,h median pooled over tasks), and 174 of 1{,}286 minutes at $M=2{,}500$. Most of the time goes into scoring. Each document needs a full 130M-parameter per-sample gradient, a rotation into the eigenbasis of 496 modules, and a contraction of its per-position hook factors against the 200 held query blocks. At $M=2{,}500$, the contraction took 374 of the 556 minutes per variant and the rotation 76. The eigenbasis (51.5\,GB) did not fit in free GPU memory beside the scoring buffers, so it stayed in host memory and was re-read for every batch.
    \item \textbf{Layerwise-project job.} Its methods avoid all three costs. A document costs one batched backward pass plus a $k\times k$ contraction per module, and the solves over $M\times M$ or $256\times256$ matrices take seconds. The price is the second pass over the corpus for the per-token terms of methods whose query operator needs the projected training gradients. At $M=2{,}500$, that pass took 107 of the 217 minutes in the stages shown.
    \item \textbf{Dense TRAK.} It is the cheapest gradient method because it computes neither curvature nor per-token terms. Its cost is one training pass with the dense projection: 12 minutes at $M=500$, 61 at $M=2{,}500$.
    \item \textbf{Trajectory methods.} Layerwise TracIn costs one pass per checkpoint and variant: 20 minutes per checkpoint on the reference cell, and 104 at $M=2{,}500$ with per-token terms. SOURCE's stages are the same. Exact TracIn and TracIn-Cos cost slightly more because they contract full gradients instead of layerwise projections.
\end{itemize}

\begin{table}[t]
\centering
\small
\caption{Per-stage wall-clock of sweep jobs on one A100-80GB, Function Bindings, $|\mathcal{F}|=50$, $|\mathcal{Q}|=200$. Parenthesized entries are linear estimates from measured no-terms jobs at $M\in\{50,250,500\}$; all other entries are medians over the seeds run on this host. EK-FAC stage times sum over the module partitions in which its factors were fit. Model loading and metric scoring are not shown.}
\label{tab:app-attr-stages}
\begin{tabular}{@{}llrrr@{}}
\toprule
 & & \multicolumn{3}{c}{Minutes} \\
\cmidrule(l){3-5}
Job & Stage & $M{=}500$ & $M{=}2{,}500$ & $M{=}2{,}500$ \\
 & & no terms & no terms & with terms \\
\midrule
\multirow{7}{*}{EK-FAC} & Kronecker factors: covariance pass & 22 & (108) & 94 \\
 & Eigendecomposition (float64) & 5 & (5) & 5 \\
 & Eigenvalue-correction pass & 18 & (89) & 74 \\
 & Scoring, CE variant & 87 & (435) & 556 \\
 & Scoring, neg.\ log-odds variant & 76 & (375) & 556 \\
 & \quad of which per-token contraction, per variant & -- & -- & 374 \\
 & \quad of which eigenbasis rotation, per variant & -- & -- & 76 \\
\midrule\multirow{4}{*}{Layerwise-Proj.} & Query gradients (projected) & 3 & (3) & 4 \\
 & Training pass, CE variant (also the Fisher) & 9 & (45) & 49 \\
 & Training pass, neg.\ log-odds variant & 9 & (46) & 57 \\
 & Second pass for per-token terms (both variants) & -- & -- & 107 \\
\midrule\multirow{2}{*}{TRAK (dense)} & Query gradients + JL & 4 & 5 & -- \\
 & Training gradients + JL (neg.\ log-odds variant) & 12 & 61 & -- \\
\midrule Layerwise TracIn & Per checkpoint, both variants ($\times$8) & 20 ($\times$8) & (88) ($\times$8) & 104 ($\times$8) \\
\bottomrule
\end{tabular}
\end{table}

\section{Full Results}
\label{app:results}

This appendix reports the numbers behind the main-text figures for every method and variant, repeats the main-text figures that show a single setting on the other settings (Section~\ref{app:res-functions}), and summarizes the single-setting score and token analyses on the other tasks (Section~\ref{app:res-other-tasks}). The tables come from the same snapshot of the sweep as the figures. Unless stated otherwise, cells are at $|\mathcal{F}|=50$ and average over seeds: three for most methods, two for the training-dynamics methods (trajectories exist for two seeds; Appendix~\ref{app:tb-finetune}), and one for BM25, which does not read the fine-tuned model.

Each method appears in two configurations:
\begin{itemize}
    \item \textbf{Document-level score} (``Doc.''): the method's own training-side loss, as drawn in Figure~\ref{fig:recall_naive}. This is cross-entropy for every method except TRAK and Layerwise TRAK, whose own recipe is the log-odds with the $Q=1-\bar p$ weight.
    \item \textbf{Neg.\ log-odds top-10} (``Top-10''): each document is scored by the mean of its ten largest per-token log-odds terms (Section~\ref{sec:topk_pooling}). A dagger ($^\dagger$) marks a method without per-token terms, which is scored with the full-document log-odds instead. TRAK and Layerwise TRAK drop $Q$ in this configuration, so that no method's top-10 score carries a document weight (Section~\ref{app:methods-trak}).
\end{itemize}
The metrics are Recall@$N$, the fraction of a query's $N$ ground-truth documents among its $N$ highest-ranked documents, and AUC, both computed per query and averaged. AUC is the probability that a random ground-truth document of the query's fact outranks a random other document: 0.5 is chance, while 1.0 means every ground-truth document ranks above every other document. On Function Bindings with Distractors we also report \emph{decoy AUC}, which ranks each ground-truth document only against the decoys written for the same query. A decoy AUC below 0.5 means that the method prefers the decoys. Where a table averages over $N$, it uses $N\in\{1,5,10,30,50\}$ on the Function Bindings tasks and $N\in\{5,10,30,50\}$ on Rule Routings. These are the values that both Figure~\ref{fig:recall_naive} and Figure~\ref{fig:recall_improved} cover; the mean is omitted (``--'') for a method missing any of them.

\subsection{Recall by Number of Documents}
\label{app:res-recall}

\begin{figure}[t]
    \centering
    \IfFileExists{arxiv/figures/appendix_figures/res_fixed_recall.pdf}{%
        \includegraphics[width=\linewidth]{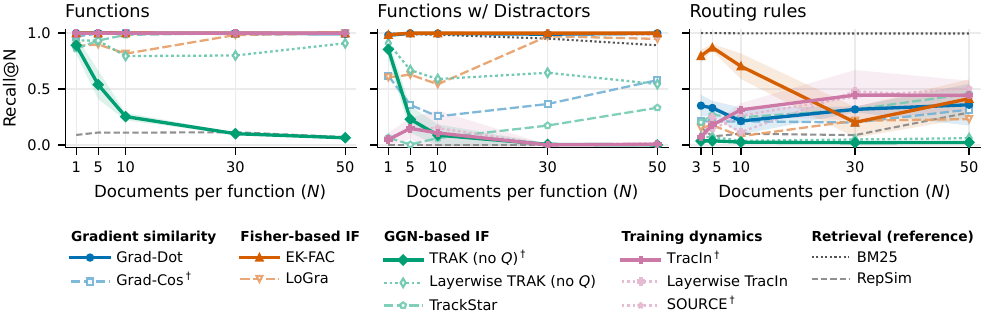}%
    }{%
        \fbox{\parbox[c][4.5cm][c]{0.95\linewidth}{\centering Upload \texttt{res\_fixed\_recall.pdf} to \texttt{\tbfigdir}.}}%
    }
    \caption{\textbf{Recall@$N$ with log-odds top-10 scoring.} The counterpart of Figure~\ref{fig:recall_naive}. Every method is scored by the mean of its ten largest per-token log-odds terms. Methods marked $^\dagger$ have no per-token terms and use the full-document log-odds. Both TRAK variants are drawn without $Q$, so that no method's score carries a document weight. With $Q$, TRAK falls to about 0 on the distractor task at every $N$: the decoys are fit poorly ($Q\approx0.17$--$0.20$) and the ground-truth documents are memorized ($Q\approx0.007$--$0.009$), so $Q$ ranks a decoy first for every query. The panel titles and method names follow the main-text figures.}
    \label{fig:app-res-fixed}
\end{figure}

Tables~\ref{tab:app-res-recall-vtok101}--\ref{tab:app-res-recall-route} give Recall@$N$ for every method at every $N$, and Figure~\ref{fig:app-res-fixed} plots the top-10 configuration as Figure~\ref{fig:recall_naive} plots the document-level one.

\begin{itemize}
    \item \textbf{Function Bindings.} At the document level, the only attribution methods above 0.5 on average are TRAK at the matched width (0.76), Layerwise TRAK (0.62) and EK-FAC (0.59). Every method except the TRAK variants reaches at least 0.92 with log-odds top-10 scoring. The TRAK variants without per-token terms keep their document-level recall; dense TRAK falls from 0.88 at $N=1$ to 0.07 at $N=50$. RepSim reaches only 0.10.
    \item \textbf{With distractors.} At the document level, dense and Layerwise TRAK are at 0.00, and every other gradient method, including EK-FAC (0.06), is below 0.2. With top-10 scoring, Grad-Dot, EK-FAC and Layerwise TracIn reach 0.99--1.00 and LoGra-Dot 0.93. TrackStar (0.13) does not recover (Section~\ref{app:res-arms}). Neither do SOURCE (0.08), exact TracIn (0.06) and Grad-Cos (0.44), which have no per-token terms and are scored on the full document; TracIn-Cos, scored the same way, recovers partly (0.72). RepSim is at 0.00.
    \item \textbf{Rule Routings.} No attribution method approaches BM25 (1.00), which is strong here because the documents of a rule share its property words with the query (Appendix~\ref{app:tb-routing}). EK-FAC is the strongest gradient method at the document level (0.19) and with top-10 scoring (0.55). No other method exceeds 0.35, which exact TracIn and TracIn-Cos reach with the full-document log-odds. RepSim, which compares the fine-tuned model's last-token states, reaches 0.14.
\end{itemize}

\begin{table}[t]
\centering
\scriptsize
\setlength{\tabcolsep}{4pt}
\caption{Recall@$N$ on \textbf{Function Bindings}, mean\,{\tiny$\pm$\,standard deviation} over seeds, at $|\mathcal{F}|=50$. \emph{Top block}: document-level scores, as in Figure~\ref{fig:recall_naive}. \emph{Bottom block}: log-odds top-10 scores, as in Figure~\ref{fig:app-res-fixed}; $^\dagger$: no per-token terms, full-document log-odds. \emph{Mean}: over $N$. A missing $\pm$ means a single seed. ``--'': not run.}
\label{tab:app-res-recall-vtok101}
\begin{tabular}{@{}lcccccc@{}}
\toprule
Method & $N{=}1$ & $N{=}5$ & $N{=}10$ & $N{=}30$ & $N{=}50$ & Mean \\
\midrule
\multicolumn{7}{@{}l}{\emph{Document-level score (variants of Figure~\ref{fig:recall_naive})}} \\
\midrule
Grad-Dot & 0.01\,{\tiny$\pm$.01} & 0.03\,{\tiny$\pm$.02} & 0.12\,{\tiny$\pm$.02} & 0.09\,{\tiny$\pm$.01} & 0.11\,{\tiny$\pm$.01} & 0.07 \\
Grad-Cos & 0.01\,{\tiny$\pm$.01} & 0.03\,{\tiny$\pm$.02} & 0.12\,{\tiny$\pm$.03} & 0.12\,{\tiny$\pm$.02} & 0.13\,{\tiny$\pm$.01} & 0.08 \\
\addlinespace[2pt]
EK-FAC & 0.70\,{\tiny$\pm$.02} & 0.74\,{\tiny$\pm$.01} & 0.67\,{\tiny$\pm$.02} & 0.37\,{\tiny$\pm$.02} & 0.45\,{\tiny$\pm$.02} & 0.59 \\
LoGra & 0.02\,{\tiny$\pm$.02} & 0.05\,{\tiny$\pm$.01} & 0.09\,{\tiny$\pm$.01} & 0.10\,{\tiny$\pm$.01} & 0.12\,{\tiny$\pm$.01} & 0.08 \\
LoGra-Dot & 0.01\,{\tiny$\pm$.02} & 0.03\,{\tiny$\pm$.01} & 0.07\,{\tiny$\pm$.01} & 0.04\,{\tiny$\pm$.01} & 0.04\,{\tiny$\pm$.01} & 0.04 \\
\addlinespace[2pt]
TRAK & 0.88\,{\tiny$\pm$.06} & 0.53\,{\tiny$\pm$.10} & 0.24\,{\tiny$\pm$.03} & 0.08\,{\tiny$\pm$.00} & 0.07\,{\tiny$\pm$.00} & 0.36 \\
TRAK ($k{=}126{,}976$) & 0.97\,{\tiny$\pm$.03} & 0.96\,{\tiny$\pm$.02} & 0.84\,{\tiny$\pm$.03} & 0.52\,{\tiny$\pm$.02} & 0.51\,{\tiny$\pm$.03} & 0.76 \\
Layerwise TRAK & 0.95\,{\tiny$\pm$.04} & 0.82\,{\tiny$\pm$.04} & 0.61\,{\tiny$\pm$.04} & 0.33\,{\tiny$\pm$.01} & 0.37\,{\tiny$\pm$.02} & 0.62 \\
TrackStar & 0.22\,{\tiny$\pm$.06} & 0.09\,{\tiny$\pm$.02} & 0.15\,{\tiny$\pm$.01} & 0.13\,{\tiny$\pm$.01} & 0.14\,{\tiny$\pm$.01} & 0.14 \\
\addlinespace[2pt]
TracIn & 0.11\,{\tiny$\pm$.01} & 0.15\,{\tiny$\pm$.02} & 0.34\,{\tiny$\pm$.11} & 0.21\,{\tiny$\pm$.09} & 0.22\,{\tiny$\pm$.02} & 0.21 \\
TracIn-Cos & 0.13\,{\tiny$\pm$.02} & 0.29\,{\tiny$\pm$.01} & 0.41\,{\tiny$\pm$.06} & 0.28\,{\tiny$\pm$.02} & 0.32\,{\tiny$\pm$.08} & 0.28 \\
Layerwise TracIn & 0.08\,{\tiny$\pm$.01} & 0.13\,{\tiny$\pm$.02} & 0.17\,{\tiny$\pm$.05} & 0.18\,{\tiny$\pm$.08} & 0.17\,{\tiny$\pm$.01} & 0.14 \\
Layerwise TracIn-Cos & 0.09\,{\tiny$\pm$.02} & 0.18\,{\tiny$\pm$.02} & 0.22\,{\tiny$\pm$.05} & 0.24\,{\tiny$\pm$.02} & 0.27\,{\tiny$\pm$.07} & 0.20 \\
SOURCE & 0.19\,{\tiny$\pm$.02} & 0.18\,{\tiny$\pm$.02} & 0.25\,{\tiny$\pm$.08} & 0.25\,{\tiny$\pm$.04} & 0.27\,{\tiny$\pm$.08} & 0.23 \\
\addlinespace[2pt]
BM25 & 1.00 & 1.00 & 1.00 & 1.00 & 1.00 & 1.00 \\
RepSim & 0.09\,{\tiny$\pm$.04} & 0.11\,{\tiny$\pm$.01} & 0.11\,{\tiny$\pm$.03} & 0.12\,{\tiny$\pm$.01} & 0.07\,{\tiny$\pm$.01} & 0.10 \\
\midrule
\multicolumn{7}{@{}l}{\emph{Neg.\ log-odds, top-10 (Figure~\ref{fig:app-res-fixed})}} \\
\midrule
Grad-Dot & 1.00\,{\tiny$\pm$.00} & 1.00\,{\tiny$\pm$.00} & 1.00\,{\tiny$\pm$.00} & 1.00\,{\tiny$\pm$.00} & 0.99\,{\tiny$\pm$.01} & 1.00 \\
Grad-Cos$^\dagger$ & 0.97\,{\tiny$\pm$.03} & 1.00\,{\tiny$\pm$.00} & 1.00\,{\tiny$\pm$.00} & 0.99\,{\tiny$\pm$.01} & 1.00\,{\tiny$\pm$.00} & 0.99 \\
\addlinespace[2pt]
EK-FAC & 1.00\,{\tiny$\pm$.00} & 1.00\,{\tiny$\pm$.00} & 1.00\,{\tiny$\pm$.00} & 1.00\,{\tiny$\pm$.00} & 1.00\,{\tiny$\pm$.00} & 1.00 \\
LoGra & 0.89\,{\tiny$\pm$.08} & 0.90\,{\tiny$\pm$.14} & 0.82\,{\tiny$\pm$.16} & 0.98\,{\tiny$\pm$.02} & 1.00\,{\tiny$\pm$.00} & 0.92 \\
LoGra-Dot & 0.97\,{\tiny$\pm$.02} & 0.99\,{\tiny$\pm$.01} & 0.96\,{\tiny$\pm$.06} & 0.96\,{\tiny$\pm$.02} & 0.99\,{\tiny$\pm$.01} & 0.98 \\
\addlinespace[2pt]
TRAK (no $Q$)$^\dagger$ & 0.89\,{\tiny$\pm$.07} & 0.54\,{\tiny$\pm$.11} & 0.25\,{\tiny$\pm$.03} & 0.10\,{\tiny$\pm$.01} & 0.07\,{\tiny$\pm$.00} & 0.37 \\
TRAK ($k{=}126{,}976$) (no $Q$)$^\dagger$ & 0.97\,{\tiny$\pm$.03} & 0.96\,{\tiny$\pm$.02} & 0.86\,{\tiny$\pm$.03} & 0.64\,{\tiny$\pm$.02} & 0.52\,{\tiny$\pm$.03} & 0.79 \\
Layerwise TRAK (no $Q$) & 0.93\,{\tiny$\pm$.05} & 0.93\,{\tiny$\pm$.05} & 0.80\,{\tiny$\pm$.13} & 0.80\,{\tiny$\pm$.02} & 0.91\,{\tiny$\pm$.05} & 0.87 \\
TrackStar & 0.86\,{\tiny$\pm$.13} & 0.93\,{\tiny$\pm$.09} & 0.98\,{\tiny$\pm$.02} & 1.00\,{\tiny$\pm$.00} & 1.00\,{\tiny$\pm$.00} & 0.96 \\
\addlinespace[2pt]
TracIn$^\dagger$ & 1.00\,{\tiny$\pm$.00} & 1.00\,{\tiny$\pm$.00} & 1.00\,{\tiny$\pm$.00} & 1.00\,{\tiny$\pm$.00} & 1.00\,{\tiny$\pm$.00} & 1.00 \\
TracIn-Cos$^\dagger$ & 1.00\,{\tiny$\pm$.00} & 1.00\,{\tiny$\pm$.00} & 1.00\,{\tiny$\pm$.00} & 1.00\,{\tiny$\pm$.00} & 1.00\,{\tiny$\pm$.00} & 1.00 \\
Layerwise TracIn & 0.99\,{\tiny$\pm$.01} & 1.00\,{\tiny$\pm$.00} & 1.00\,{\tiny$\pm$.00} & 1.00\,{\tiny$\pm$.00} & 1.00\,{\tiny$\pm$.00} & 1.00 \\
Layerwise TracIn-Cos$^\dagger$ & 0.99\,{\tiny$\pm$.01} & 1.00\,{\tiny$\pm$.00} & 1.00\,{\tiny$\pm$.00} & 1.00\,{\tiny$\pm$.00} & 1.00\,{\tiny$\pm$.00} & 1.00 \\
SOURCE$^\dagger$ & 1.00\,{\tiny$\pm$.00} & 1.00\,{\tiny$\pm$.00} & 1.00\,{\tiny$\pm$.00} & 1.00\,{\tiny$\pm$.00} & 1.00\,{\tiny$\pm$.00} & 1.00 \\
\bottomrule
\end{tabular}
\end{table}

\begin{table}[t]
\centering
\scriptsize
\setlength{\tabcolsep}{4pt}
\caption{Recall@$N$ on \textbf{Function Bindings with Distractors}. Layout as in Table~\ref{tab:app-res-recall-vtok101}. The corpus has $2|\mathcal{F}|N$ documents, one decoy per ground-truth document.}
\label{tab:app-res-recall-vtok101-distr}
\begin{tabular}{@{}lcccccc@{}}
\toprule
Method & $N{=}1$ & $N{=}5$ & $N{=}10$ & $N{=}30$ & $N{=}50$ & Mean \\
\midrule
\multicolumn{7}{@{}l}{\emph{Document-level score (variants of Figure~\ref{fig:recall_naive})}} \\
\midrule
Grad-Dot & 0.00\,{\tiny$\pm$.00} & 0.00\,{\tiny$\pm$.00} & 0.00\,{\tiny$\pm$.00} & 0.00\,{\tiny$\pm$.00} & 0.01\,{\tiny$\pm$.00} & 0.00 \\
Grad-Cos & 0.01\,{\tiny$\pm$.01} & 0.01\,{\tiny$\pm$.01} & 0.04\,{\tiny$\pm$.03} & 0.07\,{\tiny$\pm$.01} & 0.08\,{\tiny$\pm$.01} & 0.04 \\
\addlinespace[2pt]
EK-FAC & 0.07\,{\tiny$\pm$.06} & 0.02\,{\tiny$\pm$.02} & 0.01\,{\tiny$\pm$.01} & 0.03\,{\tiny$\pm$.02} & 0.18\,{\tiny$\pm$.04} & 0.06 \\
LoGra & 0.00\,{\tiny$\pm$.00} & 0.00\,{\tiny$\pm$.00} & 0.00\,{\tiny$\pm$.00} & 0.00\,{\tiny$\pm$.00} & 0.02\,{\tiny$\pm$.01} & 0.00 \\
LoGra-Dot & 0.00\,{\tiny$\pm$.00} & 0.00\,{\tiny$\pm$.00} & 0.00\,{\tiny$\pm$.00} & 0.00\,{\tiny$\pm$.00} & 0.00\,{\tiny$\pm$.00} & 0.00 \\
\addlinespace[2pt]
TRAK & 0.01\,{\tiny$\pm$.00} & 0.00\,{\tiny$\pm$.00} & 0.00\,{\tiny$\pm$.00} & 0.00\,{\tiny$\pm$.00} & 0.00\,{\tiny$\pm$.00} & 0.00 \\
TRAK ($k{=}126{,}976$) & 0.01\,{\tiny$\pm$.01} & 0.00\,{\tiny$\pm$.00} & 0.00\,{\tiny$\pm$.00} & 0.00\,{\tiny$\pm$.00} & 0.00\,{\tiny$\pm$.00} & 0.00 \\
Layerwise TRAK & 0.01\,{\tiny$\pm$.01} & 0.00\,{\tiny$\pm$.00} & 0.00\,{\tiny$\pm$.00} & 0.00\,{\tiny$\pm$.00} & 0.00\,{\tiny$\pm$.00} & 0.00 \\
TrackStar & 0.02\,{\tiny$\pm$.02} & 0.00\,{\tiny$\pm$.00} & 0.07\,{\tiny$\pm$.03} & 0.02\,{\tiny$\pm$.01} & 0.03\,{\tiny$\pm$.01} & 0.03 \\
\addlinespace[2pt]
TracIn & 0.04\,{\tiny$\pm$.06} & 0.06\,{\tiny$\pm$.02} & 0.03\,{\tiny$\pm$.01} & 0.10\,{\tiny$\pm$.06} & 0.03\,{\tiny$\pm$.01} & 0.05 \\
TracIn-Cos & 0.10\,{\tiny$\pm$.06} & 0.13\,{\tiny$\pm$.00} & 0.14\,{\tiny$\pm$.02} & 0.23\,{\tiny$\pm$.04} & 0.14\,{\tiny$\pm$.01} & 0.15 \\
Layerwise TracIn & 0.03\,{\tiny$\pm$.04} & 0.03\,{\tiny$\pm$.02} & 0.03\,{\tiny$\pm$.01} & 0.08\,{\tiny$\pm$.05} & 0.02\,{\tiny$\pm$.00} & 0.04 \\
Layerwise TracIn-Cos & 0.07\,{\tiny$\pm$.03} & 0.11\,{\tiny$\pm$.01} & 0.12\,{\tiny$\pm$.01} & 0.21\,{\tiny$\pm$.03} & 0.12\,{\tiny$\pm$.02} & 0.13 \\
SOURCE & 0.00\,{\tiny$\pm$.00} & 0.02\,{\tiny$\pm$.01} & 0.01\,{\tiny$\pm$.01} & 0.08\,{\tiny$\pm$.09} & 0.02\,{\tiny$\pm$.02} & 0.03 \\
\addlinespace[2pt]
BM25 & 1.00 & 0.99 & 0.99 & 0.95 & 0.89 & 0.96 \\
RepSim & 0.00\,{\tiny$\pm$.00} & 0.00\,{\tiny$\pm$.00} & 0.00\,{\tiny$\pm$.00} & 0.00\,{\tiny$\pm$.00} & 0.00\,{\tiny$\pm$.00} & 0.00 \\
\midrule
\multicolumn{7}{@{}l}{\emph{Neg.\ log-odds, top-10 (Figure~\ref{fig:app-res-fixed})}} \\
\midrule
Grad-Dot & 0.98\,{\tiny$\pm$.02} & 1.00\,{\tiny$\pm$.00} & 1.00\,{\tiny$\pm$.00} & 0.99\,{\tiny$\pm$.01} & 1.00\,{\tiny$\pm$.00} & 0.99 \\
Grad-Cos$^\dagger$ & 0.62\,{\tiny$\pm$.20} & 0.36\,{\tiny$\pm$.19} & 0.26\,{\tiny$\pm$.34} & 0.37\,{\tiny$\pm$.16} & 0.58\,{\tiny$\pm$.31} & 0.44 \\
\addlinespace[2pt]
EK-FAC & 0.98\,{\tiny$\pm$.02} & 1.00\,{\tiny$\pm$.00} & 1.00\,{\tiny$\pm$.00} & 1.00\,{\tiny$\pm$.00} & 1.00\,{\tiny$\pm$.00} & 1.00 \\
LoGra & 0.60\,{\tiny$\pm$.07} & 0.63\,{\tiny$\pm$.21} & 0.54\,{\tiny$\pm$.30} & 0.97\,{\tiny$\pm$.02} & 0.95\,{\tiny$\pm$.07} & 0.74 \\
LoGra-Dot & 0.96\,{\tiny$\pm$.04} & 0.95\,{\tiny$\pm$.03} & 0.91\,{\tiny$\pm$.06} & 0.91\,{\tiny$\pm$.02} & 0.94\,{\tiny$\pm$.04} & 0.93 \\
\addlinespace[2pt]
TRAK (no $Q$)$^\dagger$ & 0.85\,{\tiny$\pm$.07} & 0.23\,{\tiny$\pm$.10} & 0.09\,{\tiny$\pm$.08} & 0.01\,{\tiny$\pm$.01} & 0.00\,{\tiny$\pm$.00} & 0.24 \\
TRAK ($k{=}126{,}976$) (no $Q$)$^\dagger$ & 0.94\,{\tiny$\pm$.03} & 0.53\,{\tiny$\pm$.18} & 0.33\,{\tiny$\pm$.32} & 0.18\,{\tiny$\pm$.11} & 0.04\,{\tiny$\pm$.04} & 0.40 \\
Layerwise TRAK (no $Q$) & 0.91\,{\tiny$\pm$.09} & 0.67\,{\tiny$\pm$.16} & 0.59\,{\tiny$\pm$.25} & 0.65\,{\tiny$\pm$.07} & 0.55\,{\tiny$\pm$.30} & 0.67 \\
TrackStar & 0.07\,{\tiny$\pm$.05} & 0.00\,{\tiny$\pm$.00} & 0.06\,{\tiny$\pm$.11} & 0.17\,{\tiny$\pm$.06} & 0.33\,{\tiny$\pm$.32} & 0.13 \\
\addlinespace[2pt]
TracIn$^\dagger$ & 0.05\,{\tiny$\pm$.04} & 0.15\,{\tiny$\pm$.07} & 0.11\,{\tiny$\pm$.14} & 0.00\,{\tiny$\pm$.00} & 0.01\,{\tiny$\pm$.01} & 0.06 \\
TracIn-Cos$^\dagger$ & 0.56\,{\tiny$\pm$.33} & 0.72\,{\tiny$\pm$.04} & 0.79\,{\tiny$\pm$.22} & 0.73\,{\tiny$\pm$.30} & 0.80\,{\tiny$\pm$.24} & 0.72 \\
Layerwise TracIn & 0.98\,{\tiny$\pm$.00} & 1.00\,{\tiny$\pm$.00} & 0.99\,{\tiny$\pm$.01} & 1.00\,{\tiny$\pm$.00} & 0.99\,{\tiny$\pm$.02} & 0.99 \\
Layerwise TracIn-Cos$^\dagger$ & 0.52\,{\tiny$\pm$.30} & 0.72\,{\tiny$\pm$.06} & 0.76\,{\tiny$\pm$.27} & 0.67\,{\tiny$\pm$.30} & 0.78\,{\tiny$\pm$.26} & 0.69 \\
SOURCE$^\dagger$ & 0.06\,{\tiny$\pm$.06} & 0.14\,{\tiny$\pm$.04} & 0.15\,{\tiny$\pm$.21} & 0.00\,{\tiny$\pm$.01} & 0.01\,{\tiny$\pm$.02} & 0.08 \\
\bottomrule
\end{tabular}
\end{table}

\begin{table}[t]
\centering
\scriptsize
\setlength{\tabcolsep}{4pt}
\caption{Recall@$N$ on \textbf{Rule Routings}. Layout as in Table~\ref{tab:app-res-recall-vtok101}. The $N=1$ column lies outside the Rule Routings grid (Appendix~\ref{app:tb-routing}) and is shown for completeness; the mean is over $N\in\{5,10,30,50\}$.}
\label{tab:app-res-recall-route}
\begin{tabular}{@{}lccccccc@{}}
\toprule
Method & $N{=}1$ & $N{=}3$ & $N{=}5$ & $N{=}10$ & $N{=}30$ & $N{=}50$ & Mean \\
\midrule
\multicolumn{8}{@{}l}{\emph{Document-level score (variants of Figure~\ref{fig:recall_naive})}} \\
\midrule
Grad-Dot & 0.02\,{\tiny$\pm$.00} & 0.02\,{\tiny$\pm$.00} & 0.02\,{\tiny$\pm$.01} & 0.02\,{\tiny$\pm$.01} & 0.07\,{\tiny$\pm$.01} & 0.06\,{\tiny$\pm$.01} & 0.04 \\
Grad-Cos & 0.02\,{\tiny$\pm$.00} & 0.04\,{\tiny$\pm$.01} & 0.03\,{\tiny$\pm$.01} & 0.03\,{\tiny$\pm$.00} & 0.08\,{\tiny$\pm$.01} & 0.06\,{\tiny$\pm$.01} & 0.05 \\
\addlinespace[2pt]
EK-FAC & 0.07\,{\tiny$\pm$.07} & 0.26\,{\tiny$\pm$.04} & 0.28\,{\tiny$\pm$.01} & 0.32\,{\tiny$\pm$.01} & 0.05\,{\tiny$\pm$.02} & 0.12\,{\tiny$\pm$.01} & 0.19 \\
LoGra & 0.01\,{\tiny$\pm$.01} & 0.05\,{\tiny$\pm$.01} & 0.04\,{\tiny$\pm$.01} & 0.03\,{\tiny$\pm$.00} & 0.07\,{\tiny$\pm$.01} & 0.11\,{\tiny$\pm$.01} & 0.06 \\
LoGra-Dot & 0.01\,{\tiny$\pm$.00} & 0.02\,{\tiny$\pm$.00} & 0.02\,{\tiny$\pm$.01} & 0.02\,{\tiny$\pm$.00} & 0.04\,{\tiny$\pm$.00} & 0.06\,{\tiny$\pm$.00} & 0.03 \\
\addlinespace[2pt]
TRAK & -- & 0.03\,{\tiny$\pm$.00} & 0.03\,{\tiny$\pm$.01} & 0.02\,{\tiny$\pm$.00} & 0.02\,{\tiny$\pm$.00} & 0.02\,{\tiny$\pm$.00} & 0.03 \\
TRAK ($k{=}126{,}976$) & -- & 0.10\,{\tiny$\pm$.02} & 0.11\,{\tiny$\pm$.03} & 0.10\,{\tiny$\pm$.02} & 0.08\,{\tiny$\pm$.01} & 0.11\,{\tiny$\pm$.01} & 0.10 \\
Layerwise TRAK & 0.07\,{\tiny$\pm$.03} & 0.08\,{\tiny$\pm$.01} & 0.08\,{\tiny$\pm$.03} & 0.06\,{\tiny$\pm$.01} & 0.06\,{\tiny$\pm$.00} & 0.07\,{\tiny$\pm$.01} & 0.07 \\
TrackStar & 0.04\,{\tiny$\pm$.02} & 0.05\,{\tiny$\pm$.00} & 0.03\,{\tiny$\pm$.00} & 0.04\,{\tiny$\pm$.01} & 0.09\,{\tiny$\pm$.01} & 0.11\,{\tiny$\pm$.01} & 0.07 \\
\addlinespace[2pt]
TracIn & -- & 0.04\,{\tiny$\pm$.00} & 0.03\,{\tiny$\pm$.01} & 0.08\,{\tiny$\pm$.01} & 0.09\,{\tiny$\pm$.03} & 0.05\,{\tiny$\pm$.00} & 0.06 \\
TracIn-Cos & -- & 0.04\,{\tiny$\pm$.00} & 0.04\,{\tiny$\pm$.00} & 0.06\,{\tiny$\pm$.00} & 0.06\,{\tiny$\pm$.01} & 0.05\,{\tiny$\pm$.00} & 0.05 \\
Layerwise TracIn & 0.02\,{\tiny$\pm$.02} & 0.03\,{\tiny$\pm$.01} & 0.03\,{\tiny$\pm$.00} & 0.07\,{\tiny$\pm$.01} & 0.08\,{\tiny$\pm$.02} & 0.05\,{\tiny$\pm$.00} & 0.06 \\
Layerwise TracIn-Cos & 0.03\,{\tiny$\pm$.00} & 0.04\,{\tiny$\pm$.01} & 0.04\,{\tiny$\pm$.00} & 0.06\,{\tiny$\pm$.01} & 0.06\,{\tiny$\pm$.02} & 0.05\,{\tiny$\pm$.00} & 0.05 \\
SOURCE & 0.02\,{\tiny$\pm$.01} & 0.03\,{\tiny$\pm$.01} & 0.04\,{\tiny$\pm$.00} & 0.09\,{\tiny$\pm$.00} & 0.09\,{\tiny$\pm$.03} & 0.07\,{\tiny$\pm$.01} & 0.08 \\
\addlinespace[2pt]
BM25 & 1.00 & 1.00 & 1.00 & 1.00 & 1.00 & 1.00 & 1.00 \\
RepSim & -- & 0.16\,{\tiny$\pm$.09} & 0.08\,{\tiny$\pm$.05} & 0.10\,{\tiny$\pm$.08} & 0.09\,{\tiny$\pm$.07} & 0.29\,{\tiny$\pm$.09} & 0.14 \\
\midrule
\multicolumn{8}{@{}l}{\emph{Neg.\ log-odds, top-10 (Figure~\ref{fig:app-res-fixed})}} \\
\midrule
Grad-Dot & 0.16\,{\tiny$\pm$.09} & 0.35\,{\tiny$\pm$.08} & 0.33\,{\tiny$\pm$.05} & 0.22\,{\tiny$\pm$.10} & 0.32\,{\tiny$\pm$.02} & 0.36\,{\tiny$\pm$.16} & 0.31 \\
Grad-Cos$^\dagger$ & 0.10\,{\tiny$\pm$.06} & 0.22\,{\tiny$\pm$.06} & 0.21\,{\tiny$\pm$.06} & 0.21\,{\tiny$\pm$.04} & 0.20\,{\tiny$\pm$.04} & 0.32\,{\tiny$\pm$.09} & 0.23 \\
\addlinespace[2pt]
EK-FAC & 0.40\,{\tiny$\pm$.16} & 0.80\,{\tiny$\pm$.05} & 0.87\,{\tiny$\pm$.03} & 0.70\,{\tiny$\pm$.10} & 0.20\,{\tiny$\pm$.12} & 0.41\,{\tiny$\pm$.15} & 0.55 \\
LoGra & 0.08\,{\tiny$\pm$.06} & 0.15\,{\tiny$\pm$.06} & 0.18\,{\tiny$\pm$.13} & 0.08\,{\tiny$\pm$.05} & 0.22\,{\tiny$\pm$.06} & 0.23\,{\tiny$\pm$.15} & 0.18 \\
LoGra-Dot & 0.12\,{\tiny$\pm$.06} & 0.20\,{\tiny$\pm$.09} & 0.21\,{\tiny$\pm$.15} & 0.09\,{\tiny$\pm$.07} & 0.13\,{\tiny$\pm$.03} & 0.21\,{\tiny$\pm$.15} & 0.16 \\
\addlinespace[2pt]
TRAK (no $Q$)$^\dagger$ & -- & 0.04\,{\tiny$\pm$.01} & 0.04\,{\tiny$\pm$.01} & 0.03\,{\tiny$\pm$.00} & 0.02\,{\tiny$\pm$.00} & 0.02\,{\tiny$\pm$.00} & 0.03 \\
TRAK ($k{=}126{,}976$) (no $Q$)$^\dagger$ & -- & 0.14\,{\tiny$\pm$.02} & 0.14\,{\tiny$\pm$.03} & 0.11\,{\tiny$\pm$.03} & 0.09\,{\tiny$\pm$.01} & 0.11\,{\tiny$\pm$.02} & 0.11 \\
Layerwise TRAK (no $Q$) & 0.09\,{\tiny$\pm$.01} & 0.07\,{\tiny$\pm$.01} & 0.07\,{\tiny$\pm$.04} & 0.04\,{\tiny$\pm$.01} & 0.05\,{\tiny$\pm$.01} & 0.06\,{\tiny$\pm$.03} & 0.06 \\
TrackStar & 0.09 & 0.20\,{\tiny$\pm$.05} & 0.27\,{\tiny$\pm$.03} & 0.25\,{\tiny$\pm$.03} & 0.32\,{\tiny$\pm$.07} & 0.46\,{\tiny$\pm$.10} & 0.32 \\
\addlinespace[2pt]
TracIn$^\dagger$ & -- & 0.07\,{\tiny$\pm$.04} & 0.18\,{\tiny$\pm$.01} & 0.32\,{\tiny$\pm$.04} & 0.45\,{\tiny$\pm$.16} & 0.44\,{\tiny$\pm$.09} & 0.35 \\
TracIn-Cos$^\dagger$ & -- & 0.08\,{\tiny$\pm$.03} & 0.16\,{\tiny$\pm$.00} & 0.34\,{\tiny$\pm$.02} & 0.45\,{\tiny$\pm$.16} & 0.45\,{\tiny$\pm$.08} & 0.35 \\
Layerwise TracIn & 0.05\,{\tiny$\pm$.04} & 0.05\,{\tiny$\pm$.00} & 0.25\,{\tiny$\pm$.04} & 0.12\,{\tiny$\pm$.04} & 0.48\,{\tiny$\pm$.22} & 0.43\,{\tiny$\pm$.07} & 0.32 \\
Layerwise TracIn-Cos$^\dagger$ & 0.03\,{\tiny$\pm$.01} & 0.05\,{\tiny$\pm$.00} & 0.10\,{\tiny$\pm$.00} & 0.22\,{\tiny$\pm$.01} & 0.36\,{\tiny$\pm$.15} & 0.38\,{\tiny$\pm$.07} & 0.26 \\
SOURCE$^\dagger$ & 0.04\,{\tiny$\pm$.02} & 0.07\,{\tiny$\pm$.02} & 0.14\,{\tiny$\pm$.01} & 0.26\,{\tiny$\pm$.03} & 0.43\,{\tiny$\pm$.15} & 0.51\,{\tiny$\pm$.05} & 0.34 \\
\bottomrule
\end{tabular}
\end{table}

\subsection{AUC and Decoy AUC}
\label{app:res-auc}

Table~\ref{tab:app-res-auc} gives the AUC averaged over $N$. AUC does not depend on the size of the corpus, so it separates a method that still ranks the ground-truth documents above most others from one that ranks them at random. Two patterns stand out.
\begin{itemize}
    \item \textbf{Low recall with high AUC.} Several document-level scores rank the ground truth well without putting it at the top. Dense TRAK has AUC 0.80 on Function Bindings with Recall@$N$ 0.36, and Layerwise TracIn 0.62 with 0.14. The top-10 score raises the AUC of every method with per-token terms on Function Bindings to at least 0.99.
    \item \textbf{Decoy AUC.} On the distractor task, every document-level gradient score has decoy AUC between 0.30 and 0.50, so it separates a fact's documents from their decoys at close to chance or worse. With top-10 scoring, every method with per-token terms except TrackStar reaches 1.00. TrackStar reaches 0.51, and the full-document log-odds of SOURCE and exact TracIn 0.15 and 0.13. RepSim has decoy AUC 0.00, since a decoy is written to resemble its fact's query.
\end{itemize}

\begin{table}[t]
\centering
\scriptsize
\setlength{\tabcolsep}{4pt}
\caption{AUC averaged over $N$ at $|\mathcal{F}|=50$, for the document-level score (Doc.) and the log-odds top-10 score (Top-10). The last two columns are decoy AUC on Function Bindings with Distractors: each ground-truth document ranked against the decoys of its own query. $^\dagger$: no per-token terms; the Top-10 column uses the full-document log-odds. TRAK's Doc.\ columns include $Q$ and its Top-10 columns do not. The retrieval references have one configuration.}
\label{tab:app-res-auc}
\resizebox{\linewidth}{!}{%
\begin{tabular}{@{}lcccccccc@{}}
\toprule
 & \multicolumn{2}{c}{Function Bindings} & \multicolumn{2}{c}{w/ Distractors} & \multicolumn{2}{c}{Rule Routings} & \multicolumn{2}{c}{w/ Distractors, decoy AUC} \\
\cmidrule(lr){2-3}\cmidrule(lr){4-5}\cmidrule(lr){6-7}\cmidrule(lr){8-9}
Method & Doc. & Top-10 & Doc. & Top-10 & Doc. & Top-10 & Doc. & Top-10 \\
\midrule
Grad-Dot & 0.46 & 1.00 & 0.39 & 1.00 & 0.50 & 0.85 & 0.44 & 1.00 \\
Grad-Cos$^\dagger$ & 0.46 & 1.00 & 0.31 & 0.91 & 0.50 & 0.71 & 0.30 & 0.63 \\
\addlinespace[2pt]EK-FAC & 0.69 & 1.00 & 0.53 & 1.00 & 0.52 & 0.95 & 0.45 & 1.00 \\
LoGra & 0.52 & 0.99 & 0.50 & 0.98 & 0.51 & 0.79 & 0.50 & 1.00 \\
LoGra-Dot & 0.48 & 1.00 & 0.44 & 1.00 & 0.50 & 0.78 & 0.48 & 1.00 \\
\addlinespace[2pt]TRAK$^\dagger$ & 0.80 & 0.80 & 0.64 & 0.70 & 0.51 & 0.52 & 0.42 & 0.68 \\
TRAK ($k{=}126{,}976$)$^\dagger$ & 0.92 & 0.92 & 0.72 & 0.80 & 0.54 & 0.54 & 0.39 & 0.74 \\
Layerwise TRAK & 0.89 & 0.99 & 0.68 & 0.98 & 0.53 & 0.57 & 0.40 & 1.00 \\
TrackStar & 0.55 & 0.99 & 0.50 & 0.85 & 0.51 & 0.81 & 0.49 & 0.51 \\
\addlinespace[2pt]TracIn$^\dagger$ & 0.66 & 1.00 & 0.52 & 0.86 & 0.54 & 0.80 & 0.42 & 0.13 \\
TracIn-Cos$^\dagger$ & 0.67 & 1.00 & 0.49 & 0.99 & 0.54 & 0.78 & 0.44 & 0.74 \\
Layerwise TracIn & 0.62 & 1.00 & 0.54 & 1.00 & 0.54 & 0.87 & 0.43 & 1.00 \\
Layerwise TracIn-Cos$^\dagger$ & 0.63 & 1.00 & 0.52 & 0.98 & 0.53 & 0.75 & 0.45 & 0.71 \\
SOURCE$^\dagger$ & 0.67 & 1.00 & 0.56 & 0.85 & 0.55 & 0.79 & 0.41 & 0.15 \\
\addlinespace[2pt]BM25 & 1.00 & -- & 1.00 & -- & 1.00 & -- & 0.99 & -- \\
RepSim & 0.62 & -- & 0.33 & -- & 0.68 & -- & 0.00 & -- \\
\bottomrule
\end{tabular}
}
\end{table}

\subsection{Every Scoring Variant}
\label{app:res-arms}

The per-token terms of a method give many document scores at no extra cost (Section~\ref{sec:topk_pooling}). Table~\ref{tab:app-res-arms} reports Recall@$N$ for each of them. The CE columns are the full-document mean and sum and the top-$k$ means. The log-odds columns are:
\begin{itemize}
    \item the full-document mean, which is the gradient of $L_{\mathrm{lo}}$;
    \item the full-document sum;
    \item the mean weighted by $Q=1-\bar p$, which is TRAK's recipe applied to every method;
    \item the top-$k$ means.
\end{itemize}
Four observations follow.
\begin{itemize}
    \item \textbf{The loss matters before pooling.} On the two Function Bindings tasks, top-$k$ pooling of the CE terms recovers EK-FAC (from 0.59 to 0.89, and from 0.06 to 0.87) but leaves every other method at 0.22 or below. Switching to log-odds terms lifts every method on Function Bindings, even without pooling.
    \item \textbf{The full-document mean favors short documents.} With distractors, the full-document mean fails under either loss (Grad-Dot 0.00 and EK-FAC 0.06 under CE, 0.05 and 0.18 under log-odds), while the log-odds sum succeeds (0.85 and 0.99). The mean divides each document's score by its length, $n_d$, and the 12-token decoys are an order of magnitude shorter than the fact documents. Top-$k$ pooling does not depend on length when $k$ is below the shortest document, and it gives the best score on most rows.
    \item \textbf{$Q$ collapses every method on distractors.} Weighting by $1-\bar p$ drops every method to 0.02 or below with distractors, because the poorly fit decoys receive the largest weights. On the clean tasks, $Q$ costs between 0.00 and 0.10.
    \item \textbf{TrackStar is the exception with distractors.} Its top-10 score (0.13) is below its full-document mean (0.18). Its score carries its own document weight, the inverse whitened norm $1/\lVert R^{-1/2}\phi\rVert$ (Section~\ref{app:methods-trackstar}), which top-$k$ pooling does not remove. We expect that weight to be large for short documents, which would explain the result, but we have not verified this directly.
\end{itemize}

\begin{table}[t]
\centering
\scriptsize
\setlength{\tabcolsep}{4pt}
\caption{Recall@$N$ of every scoring variant, averaged over $N$ at $|\mathcal{F}|=50$, for the methods with per-token terms. Top-$k$ averages the $k$ largest terms under either loss (Section~\ref{sec:topk_pooling}). ``mean${}\times Q$'' multiplies the full-document log-odds by $1-\bar p$. For Layerwise TRAK, every column except that one is unweighted. TrackStar has no sum columns, because its normalized score is the same under the sum and the mean. Bold: best in the row, within 0.005.}
\label{tab:app-res-arms}
\resizebox{\linewidth}{!}{%
\begin{tabular}{@{}llccccccccc@{}}
\toprule
 & & \multicolumn{4}{c}{Cross-entropy terms} & \multicolumn{5}{c}{Neg.\ log-odds terms} \\
\cmidrule(lr){3-6}\cmidrule(l){7-11}
Task & Method & mean & sum & top-3 & top-10 & mean & sum & mean${}\times Q$ & top-3 & top-10 \\
\midrule
\multirow{7}{*}{\parbox{1.6cm}{\raggedright Function Bindings}} & Grad-Dot & 0.07 & 0.07 & 0.14 & 0.12 & 0.99 & 0.99 & 0.98 & \textbf{1.00} & \textbf{1.00} \\
 & EK-FAC & 0.59 & 0.59 & 0.87 & 0.89 & \textbf{1.00} & \textbf{1.00} & \textbf{1.00} & \textbf{1.00} & \textbf{1.00} \\
 & LoGra & 0.08 & 0.08 & 0.12 & 0.09 & 0.89 & 0.89 & 0.86 & \textbf{0.93} & 0.92 \\
 & LoGra-Dot & 0.04 & 0.04 & 0.05 & 0.06 & 0.93 & 0.93 & 0.90 & 0.97 & \textbf{0.98} \\
 & Layerwise TRAK (no $Q$) & 0.04 & 0.04 & 0.03 & 0.04 & 0.65 & 0.64 & 0.62 & 0.84 & \textbf{0.87} \\
 & TrackStar & 0.14 & -- & 0.21 & 0.22 & 0.91 & -- & 0.89 & 0.95 & \textbf{0.96} \\
 & Layerwise TracIn & 0.14 & 0.15 & 0.16 & 0.14 & \textbf{1.00} & \textbf{1.00} & 0.97 & \textbf{0.99} & \textbf{1.00} \\
\midrule
\multirow{7}{*}{\parbox{1.6cm}{\raggedright w/ Distractors}} & Grad-Dot & 0.00 & 0.06 & 0.13 & 0.14 & 0.05 & 0.85 & 0.00 & \textbf{0.99} & \textbf{0.99} \\
 & EK-FAC & 0.06 & 0.44 & 0.83 & 0.87 & 0.18 & 0.99 & 0.01 & \textbf{0.99} & \textbf{1.00} \\
 & LoGra & 0.00 & 0.08 & 0.15 & 0.11 & 0.03 & 0.60 & 0.00 & \textbf{0.75} & 0.74 \\
 & LoGra-Dot & 0.00 & 0.04 & 0.06 & 0.07 & 0.06 & 0.64 & 0.01 & \textbf{0.94} & \textbf{0.93} \\
 & Layerwise TRAK (no $Q$) & 0.00 & 0.01 & 0.03 & 0.03 & 0.33 & 0.47 & 0.00 & 0.60 & \textbf{0.67} \\
 & TrackStar & 0.03 & -- & 0.04 & 0.05 & \textbf{0.18} & -- & 0.02 & 0.07 & 0.13 \\
 & Layerwise TracIn & 0.04 & 0.12 & 0.18 & 0.14 & 0.07 & 0.93 & 0.00 & 0.98 & \textbf{0.99} \\
\midrule
\multirow{7}{*}{\parbox{1.6cm}{\raggedright Rule Routings}} & Grad-Dot & 0.04 & 0.04 & 0.04 & 0.05 & 0.22 & 0.22 & 0.17 & \textbf{0.31} & \textbf{0.31} \\
 & EK-FAC & 0.19 & 0.19 & 0.30 & 0.35 & 0.48 & 0.48 & 0.44 & 0.47 & \textbf{0.55} \\
 & LoGra & 0.06 & 0.06 & 0.06 & 0.05 & 0.17 & 0.17 & 0.15 & \textbf{0.19} & 0.18 \\
 & LoGra-Dot & 0.03 & 0.03 & 0.03 & 0.04 & 0.13 & 0.13 & 0.10 & \textbf{0.17} & 0.16 \\
 & Layerwise TRAK (no $Q$) & 0.04 & 0.04 & 0.04 & 0.05 & \textbf{0.07} & \textbf{0.07} & 0.07 & 0.05 & 0.06 \\
 & TrackStar & 0.07 & -- & 0.13 & 0.13 & 0.18 & -- & 0.14 & 0.31 & \textbf{0.32} \\
 & Layerwise TracIn & 0.06 & 0.06 & 0.07 & 0.07 & 0.26 & 0.26 & 0.16 & 0.31 & \textbf{0.32} \\
\bottomrule
\end{tabular}
}
\end{table}

\subsection{Score Distributions and Token Types on the Other Tasks}
\label{app:res-other-tasks}

Figure~\ref{fig:attr_histogram} shows a single cell of Function Bindings with Distractors, and Figure~\ref{fig:token_analysis} shows a single Rule Routings setting. Both can be explored on the other tasks online: the score distributions for every task, method and model seed on our \href{https://catt.baulab.info/\#histograms}{project page}, and the per-token attributions of each method for any query and training document in our \href{https://huggingface.co/spaces/catt-bench/catt-demo}{demo}. Below we summarize both for Grad-Dot on the other tasks.

\begin{itemize}
    \item \textbf{Score distributions} (same cell as Figure~\ref{fig:attr_histogram}: $|\mathcal{F}|=50$, $N=50$, seed 1001). On Function Bindings, changing the loss alone separates the ground-truth documents from the rest: Recall@$N$ rises from 0.11 to 1.00 with neg.\ log-odds over all tokens and stays at 1.00 with top-10 selection, while top-10 selection under cross-entropy reaches 0.27. On Rule Routings, each change helps but neither separates the two groups. Recall@$N$ is 0.07 with cross-entropy over all tokens, 0.39 with neg.\ log-odds, 0.10 with cross-entropy top-10, and 0.47 with both changes. With neg.\ log-odds top-10 scores, Rule Routings is also the task on which most attribution methods have their lowest recall (Section~\ref{app:res-recall}); the exceptions, TrackStar, TracIn and SOURCE, score lower still on Function Bindings with distractors. With document-level scores, every attribution method except TracIn-Cos and Layerwise TracIn-Cos has its lowest recall on Function Bindings with distractors.
    \item \textbf{Token types}. On the Function Bindings tasks, each position is one of six types, taken in this order: the answer token $\langle V\rangle$, the function name $\langle B\rangle$, integers, other tokens inside code, other tokens containing a letter (prose), and the rest (markup and punctuation). We measure each type's share of a document's ten highest-scoring positions relative to its share of all positions, pooled over every (query, document) pair of the $|\mathcal{F}|=50$ cells, so that 1$\times$ is chance. Under cross-entropy, no type is strongly enriched in the top ten positions: the answer token ranges from 0.2$\times$ to 1.3$\times$ its baseline rate on Function Bindings and from 0.9$\times$ to 1.2$\times$ with distractors. Under neg.\ log-odds, the answer token is enriched 5.1--11$\times$ on Function Bindings and 2.9--6.6$\times$ with distractors, and no other type exceeds 1.4$\times$. This matches Rule Routings, where neg.\ log-odds enriches the outcome tokens (Figure~\ref{fig:token_analysis}).
\end{itemize}

\subsection{Number of Facts}
\label{app:res-functions}

\begin{figure}[t]
    \centering
    \IfFileExists{arxiv/figures/appendix_figures/res_recall_functions.pdf}{%
        \includegraphics[width=\linewidth]{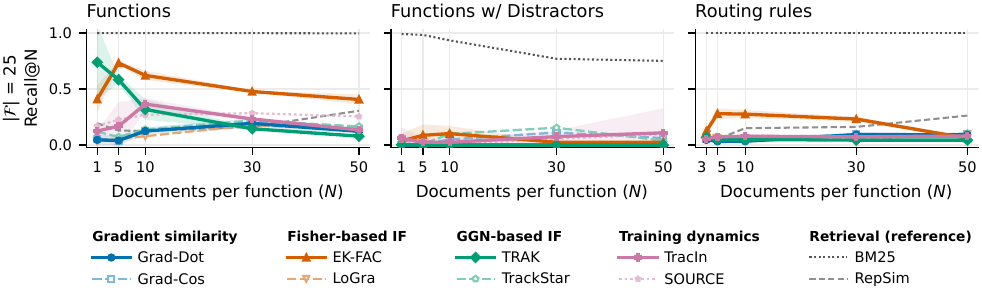}%
    }{%
        \fbox{\parbox[c][4.5cm][c]{0.95\linewidth}{\centering Upload \texttt{res\_recall\_functions.pdf} to \texttt{\tbfigdir}.}}%
    }
    \caption{\textbf{Recall@$N$ at $|\mathcal{F}|=25$.} The counterpart of Figure~\ref{fig:recall_naive}, which is at $|\mathcal{F}|=50$, with the same methods and document-level scores. The panel titles and method names follow the main-text figures.}
    \label{fig:app-res-functions}
\end{figure}

The main text fixes $|\mathcal{F}|=50$. Its conclusions do not depend on that choice: across $|\mathcal{F}|\in\{25,50,100\}$, the same methods succeed and fail on the same tasks, and recall moves much less with the number of facts than with the number of documents per fact $N$ or with distractors. Table~\ref{tab:app-res-functions} compares the three sizes at $N=10$, the document count that the $|\mathcal{F}|=100$ grid covers for the most methods, and Figure~\ref{fig:app-res-functions} repeats Figure~\ref{fig:recall_naive} at $|\mathcal{F}|=25$. The main-text trends hold at every size we ran:
\begin{itemize}
    \item \textbf{TRAK collapses as $N$ grows} on Function Bindings: from 0.74 at $N=1$ to 0.08 at $N=50$ at $|\mathcal{F}|=25$ (0.88 to 0.07 at $|\mathcal{F}|=50$), and from 0.90 at $N=1$ to 0.16 at $N=10$ at $|\mathcal{F}|=100$. EK-FAC degrades less and is the strongest attribution method at $|\mathcal{F}|=25$ as at $|\mathcal{F}|=50$ (mean recall 0.53, against 0.59).
    \item \textbf{Distractors defeat document-level scores.} With distractors, no document-level score in Table~\ref{tab:app-res-functions} exceeds 0.10 at any $|\mathcal{F}|$, and at $|\mathcal{F}|=25$ none in Figure~\ref{fig:app-res-functions} exceeds 0.16 at any $N$.
    \item \textbf{Rule Routings stays hard for gradient methods, while BM25 does not.} At $|\mathcal{F}|=25$, no attribution method exceeds 0.3 at any $N$, while BM25 is at 1.00, as at $|\mathcal{F}|=50$.
    \item \textbf{Top-10 log-odds scoring recovers Function Bindings at every size.} Grad-Dot, EK-FAC and Layerwise TracIn reach at least 0.99 on both Function Bindings tasks at every $|\mathcal{F}|$.
\end{itemize}

What changes with $|\mathcal{F}|$ is a gradual decline for some methods. EK-FAC's document-level recall on Function Bindings falls from 0.67 at $|\mathcal{F}|=50$ to 0.22 at $|\mathcal{F}|=100$. LoGra's top-10 recall falls as facts are added, most with distractors (0.90, 0.54, 0.52). On Rule Routings, most methods lose top-10 recall as rules are added; EK-FAC's falls from 0.78 to 0.70 to 0.56. TrackStar is the exception, at 0.25, 0.25 and 0.26. The $|\mathcal{F}|=100$ grid covers mostly $N\le 10$, and at $N=10$ its column is incomplete: on Function Bindings it has no TrackStar and no top-10 score for Layerwise TRAK, and it has no BM25 on Function Bindings or Rule Routings.

\begin{table}[t]
\centering
\scriptsize
\setlength{\tabcolsep}{4pt}
\caption{Recall@$N$ at $N=10$ for $|\mathcal{F}|\in\{25,50,100\}$, mean over seeds. Doc.: document-level score; Top-10: log-odds top-10 ($^\dagger$: no per-token terms, full-document log-odds; TRAK's Top-10 column without $Q$). ``--'': not run.}
\label{tab:app-res-functions}
\begin{tabular}{@{}llcccccc@{}}
\toprule
 & & \multicolumn{2}{c}{$|\mathcal{F}|{=}25$} & \multicolumn{2}{c}{$|\mathcal{F}|{=}50$} & \multicolumn{2}{c}{$|\mathcal{F}|{=}100$} \\
\cmidrule(lr){3-4}\cmidrule(lr){5-6}\cmidrule(lr){7-8}
Task & Method & Doc. & Top-10 & Doc. & Top-10 & Doc. & Top-10 \\
\midrule
\multirow{9}{*}{\parbox{1.6cm}{\raggedright Function Bindings}} & Grad-Dot & 0.13 & 1.00 & 0.12 & 1.00 & 0.12 & 0.99 \\
 & EK-FAC & 0.62 & 1.00 & 0.67 & 1.00 & 0.22 & 1.00 \\
 & LoGra & 0.08 & 0.94 & 0.09 & 0.82 & 0.11 & 0.78 \\
 & TRAK$^\dagger$ & 0.32 & 0.32 & 0.24 & 0.25 & 0.16 & 0.18 \\
 & Layerwise TRAK & 0.56 & 0.77 & 0.61 & 0.80 & 0.61 & -- \\
 & TrackStar & 0.14 & 1.00 & 0.15 & 0.98 & -- & -- \\
 & Layerwise TracIn & 0.23 & 1.00 & 0.17 & 1.00 & 0.16 & 1.00 \\
 & SOURCE$^\dagger$ & 0.27 & 1.00 & 0.25 & 1.00 & 0.27 & 1.00 \\
 & BM25 & 1.00 & -- & 1.00 & -- & -- & -- \\
\midrule
\multirow{9}{*}{\parbox{1.6cm}{\raggedright w/ Distractors}} & Grad-Dot & 0.00 & 1.00 & 0.00 & 1.00 & 0.00 & 0.99 \\
 & EK-FAC & 0.10 & 1.00 & 0.01 & 1.00 & 0.01 & 1.00 \\
 & LoGra & 0.00 & 0.90 & 0.00 & 0.54 & 0.00 & 0.52 \\
 & TRAK$^\dagger$ & 0.00 & 0.25 & 0.00 & 0.09 & 0.00 & 0.03 \\
 & Layerwise TRAK & 0.00 & 0.76 & 0.00 & 0.59 & 0.00 & 0.40 \\
 & TrackStar & 0.10 & 0.01 & 0.07 & 0.06 & 0.07 & 0.03 \\
 & Layerwise TracIn & 0.02 & 1.00 & 0.03 & 0.99 & 0.04 & 1.00 \\
 & SOURCE$^\dagger$ & 0.01 & 0.06 & 0.01 & 0.15 & 0.02 & 0.09 \\
 & BM25 & 0.94 & -- & 0.99 & -- & 0.99 & -- \\
\midrule
\multirow{9}{*}{\parbox{1.6cm}{\raggedright Rule Routings}} & Grad-Dot & 0.03 & 0.35 & 0.02 & 0.22 & 0.02 & 0.15 \\
 & EK-FAC & 0.28 & 0.78 & 0.32 & 0.70 & 0.25 & 0.56 \\
 & LoGra & 0.06 & 0.20 & 0.03 & 0.08 & 0.04 & 0.05 \\
 & TRAK$^\dagger$ & 0.05 & 0.05 & 0.02 & 0.03 & 0.01 & 0.02 \\
 & Layerwise TRAK & 0.10 & 0.08 & 0.06 & 0.04 & 0.05 & 0.02 \\
 & TrackStar & 0.05 & 0.25 & 0.04 & 0.25 & 0.04 & 0.26 \\
 & Layerwise TracIn & 0.07 & 0.32 & 0.07 & 0.12 & 0.04 & 0.19 \\
 & SOURCE$^\dagger$ & 0.08 & 0.31 & 0.09 & 0.26 & 0.06 & 0.24 \\
 & BM25 & 1.00 & -- & 1.00 & -- & -- & -- \\
\bottomrule
\end{tabular}
\end{table}

\section{FTRACE-TREX}
\label{app:ftrace}

This appendix details the external evaluation of Figure~\ref{fig:ftrace_generalization}. We follow the FTRACE-TREX setup of \citet{akyurek2022towards} and the remove-and-retrain protocol of \citet{park2023trak}, with the departures listed at the end.

\paragraph{Data and model.} The training corpus consists of 1{,}560{,}453 masked examples drawn from 484{,}408 TREx abstract sentences, and the queries are 31{,}479 cloze-style facts, each annotated with its \emph{proponents}: the training examples that state the fact. We fine-tune mT5-small (300M parameters) as a sequence-to-sequence model for 60{,}000 steps (Adafactor, constant learning rate $10^{-3}$, batch size 128, inputs truncated to 256 tokens and targets to 32; seed 1001). Top-1 accuracy with beam search rises from 10.7\% for the pretrained model to 33.7\%. Following \citet{park2023trak}, the \emph{novel} facts are the 8{,}437 queries that the pretrained model gets wrong and the fine-tuned model gets right at top-1. All attribution scores are computed on this one fine-tuned model.

\paragraph{Scoring.} Each FTRACE-TREX target frames its answer with format tokens: the sentinel \texttt{<extra\_id\_0>} before it and the end-of-sequence token \texttt{</s>} after. They occur in every target and the model predicts them almost perfectly, so we drop them from the loss on both the query and the training-document side and score only the answer tokens; the model still reads the full input and target. The per-position gradients are summed over the encoder positions (input tokens) and the decoder positions (answer tokens). We compare three methods, each as published and with our fixes:
\begin{itemize}
    \item \textbf{TRAK} applies the published dense Rademacher JL projection to all 300M parameters at $k=4{,}000$ and multiplies by $Q_d=1-\bar p_d$, where $\bar p_d$ is the mean probability of the document's answer tokens. With $n\gg k$ we solve it in the primal form $\phi_d^{\mathsf T}(\Phi^{\mathsf T}\Phi+\lambda I)^{-1}\phi_q$, which equals Eq.~\eqref{eq:app-attr-trak} before the $Q_d$ factor, with $\lambda=10^{-6}\operatorname{tr}(\Phi^{\mathsf T}\Phi)/k$. \textbf{TRAK (negative log-odds)} is the same score without $Q_d$. A dense projection does not factor over positions, so it gives no per-position terms and TRAK has no top-10 variant.
    \item \textbf{EK-FAC} preconditions with the cross-entropy EK-FAC of the fine-tuned model, damped by $0.01$ times each module's mean eigenvalue (the best of $10^{-5}$ to $1$ on 200 held-out novel facts). As published, it scores the training document's cross-entropy summed over its answer tokens; \textbf{EK-FAC (negative log-odds)} scores the negative log-odds instead.
    \item \textbf{Grad-Dot} is the same pair of scores without the preconditioner.
\end{itemize}
The \emph{top 10 tokens} variants score a document by the mean of its 10 largest per-position terms over input and answer positions. BM25 (BM25+) ranks documents by whitespace tokens of the input and target, with the mask token removed.

\paragraph{Remove-and-retrain.} We draw 50 novel facts (seed 0) from those the fine-tuned model already answers correctly at step 30{,}000, so that none is a late fluctuation. Their proponents total 11{,}522 training examples. Each method ranks the full training corpus for each fact. To give every method exactly the same budget $B$, we remove documents round-robin: each fact's next-ranked document in turn, skipping documents already removed, until $B$ documents are gone. We use $B\in\{2{,}000,\,5{,}000,\,11{,}522\}$; the largest budget equals the size of the proponent union. For each removal set, we fine-tune mT5-small from the pretrained checkpoint with the same recipe at eight seeds (2001--2008). We also retrain a no-removal control at the same seeds, which reaches 80\% to 92\% top-1 on the 50 facts (86\% on average). The reported drop is control accuracy minus retrained accuracy, paired by seed, averaged over the seeds, $\pm$ one standard error. BM25 and the ground-truth removal sets do not depend on the loss, so we reuse their earlier retrains against the same controls: two seeds at $B=2{,}000$ and $5{,}000$ (one for the ground truth at $2{,}000$) and three at $11{,}522$.

\begin{table}[t]
\centering
\small
\setlength{\tabcolsep}{3pt}
\caption{FTRACE-TREX remove-and-retrain: drop in top-1 accuracy (percentage points) on the 50 facts at each removal budget, mean over eight retraining seeds $\pm$ one standard error. \emph{Score}: the training-document score as published (TRAK's log-odds with $Q_d$, cross-entropy for EK-FAC and Grad-Dot) or the negative log-odds; \emph{Pooling}: summed over all positions or the mean of the top 10 per-position terms. The last two columns are at $B=11{,}522$. \emph{Prop.\ removed}: mean fraction of each fact's proponents in the removal set. \emph{Same-rel.\ drop}: the accuracy drop on the other 5{,}797 novel facts that share a relation with one of the 50. On the 2{,}590 novel facts in other relations, every set moves accuracy by at most 4 points. BM25 and the ground truth use two seeds at $B=2{,}000$ and $5{,}000$ and three at $11{,}522$. $^\dagger$One seed.}
\label{tab:app-ftrace-retrain}
\begin{tabular}{@{}lllccccc@{}}
\toprule
 & & & \multicolumn{3}{c}{Drop on the 50 facts} & \multicolumn{2}{c}{At $B=11{,}522$} \\
\cmidrule(lr){4-6}\cmidrule(l){7-8}
Method & Score & Pooling & $B=2{,}000$ & $5{,}000$ & $11{,}522$ & Prop.\ removed & Same-rel.\ drop \\
\midrule
TRAK & Published & All & $9\pm3$ & $10\pm2$ & $8\pm2$ & 13\% & 4.2 \\
 & Neg.\ log-odds & All & $13\pm3$ & $28\pm2$ & $40\pm4$ & 24\% & 11.7 \\
\midrule
EK-FAC & Published & All & $13\pm1$ & $25\pm2$ & $30\pm2$ & 17\% & 10.9 \\
 & Published & Top 10 & $14\pm2$ & $19\pm3$ & $34\pm2$ & 17\% & 11.9 \\
 & Neg.\ log-odds & All & $18\pm2$ & $28\pm2$ & $44\pm2$ & 27\% & 15.3 \\
 & Neg.\ log-odds & Top 10 & $\mathbf{20\pm3}$ & $\mathbf{32\pm3}$ & $\mathbf{46\pm3}$ & 28\% & 13.9 \\
\midrule
Grad-Dot & Published & All & $3\pm2$ & $7\pm2$ & $7\pm3$ & 8\% & 3.1 \\
 & Published & Top 10 & $4\pm1$ & $6\pm2$ & $12\pm4$ & 10\% & 5.5 \\
 & Neg.\ log-odds & All & $8\pm2$ & $22\pm2$ & $22\pm3$ & 17\% & 8.4 \\
 & Neg.\ log-odds & Top 10 & $11\pm1$ & $16\pm2$ & $23\pm2$ & 18\% & 7.7 \\
\midrule
\multicolumn{3}{@{}l}{BM25} & $3\pm1$ & $5\pm1$ & $15\pm5$ & 76\% & 2.4 \\
\multicolumn{3}{@{}l}{Ground-truth proponents} & $4^\dagger$ & $4\pm4$ & $9\pm3$ & 100\% & 1.1 \\
\bottomrule
\end{tabular}
\end{table}

\paragraph{Retraining results.} Table~\ref{tab:app-ftrace-retrain} gives the values plotted in Figure~\ref{fig:ftrace_generalization}, together with the remaining variants of each method. Four observations qualify the main-text result:
\begin{itemize}
    \item \textbf{The loss change helps every method; top-10 selection is roughly neutral.} Paired by seed at $B=11{,}522$, negative log-odds adds $32\pm3$ points to TRAK, $15\pm1$ to EK-FAC, and $16\pm3$ to Grad-Dot. Under negative log-odds, keeping only the top 10 tokens adds 2 to 4 points to EK-FAC at every budget and changes Grad-Dot by $-6$ to $+3$.
    \item \textbf{Removing the annotated proponents matters little.} Removing every ground-truth proponent drops accuracy on the 50 facts by only 9 points. Removing BM25's selection, which contains 76\% of the proponents, drops it by 15 points, less than removing the selections of TRAK, EK-FAC, or Grad-Dot under negative log-odds. \citet{park2023trak} found the same on FTRACE-TREX: BM25 retrieves the proponents far better than TRAK, yet removing TRAK's top-ranked documents lowers accuracy more. The documents that matter to the retrained model are largely not the annotated ones.
    \item \textbf{Larger accuracy drops come with forgetting of collateral facts.} The negative-log-odds rankings also lower accuracy on other facts of the same relations, by 8 (Grad-Dot) to 15 (EK-FAC) points. Their highest-ranked documents include ones that teach a relation's answer pattern, not only one fact's evidence. The drop on the 50 facts is still about three times this collateral, and facts in other relations are unaffected.
\end{itemize}

\paragraph{Departures from \citet{park2023trak}.}
\begin{itemize}
    \item \textbf{Ensemble.} In our results, TRAK uses one model rather than an ensemble of $M=10$ trained on random subsets.
    \item \textbf{Answer tokens only.} Every loss excludes the target's format tokens on both the query and the training-document side (see \emph{Scoring}).
    \item \textbf{Baseline for the drop.} We measure the drop against a retrained no-removal control rather than against the original model taken as 100\% accurate. A model retrained at a new seed answers only about 86\% of facts selected on seed 1001, and the 100\% baseline would count that shortfall as removal damage.
    \item \textbf{Removal budget.} Every method removes the same number of documents, rather than the union of each fact's top 500, whose size varies by method (4{,}800 to 18{,}200 documents in our runs).
\end{itemize}

\end{document}